%% file: main.tex
\documentclass[lettersize,journal]{IEEEtran}

\usepackage{amsmath,amsfonts,amssymb}
\usepackage{array}
\usepackage{textcomp}
\usepackage{url}
\usepackage{verbatim}
\usepackage{graphicx}
\usepackage{balance}

\usepackage{xcolor}
\usepackage[colorlinks=true, urlcolor=blue, linkcolor=blue, citecolor=blue]{hyperref}

\usepackage[caption=false,font=normalsize,labelfont=sf,textfont=sf]{subfig}
\usepackage{caption}
\usepackage{subcaption}

\usepackage{cuted}   % provides strip environment (spans both columns)
\usepackage{capt-of} % provides \captionof

\usepackage{algorithm}
\usepackage{algpseudocode}
\usepackage{tabularx}
\usepackage{booktabs}
\usepackage{multirow}
\usepackage{enumitem}
\usepackage{tcolorbox}
\tcbuselibrary{listings, breakable}
\usepackage{xcolor}
\usepackage{mdframed}
\usepackage{cite}


\begin{document}

\title{Mind-of-Director: Multi-modal Agent-Driven Film Previsualization via Collaborative Decision-Making}

\author{Shufeng Nan, Mengtian Li*, Sixiao Zheng, Yuwei Lu, Han Zhang, Yanwei Fu*%
\thanks{Shufeng Nan and Sixiao Zheng are with Fudan University, Shanghai, China.
E-mail: 25210980158@m.fudan.edu.cn; sxzheng18@fudan.edu.cn.}
\thanks{Han Zhang, and Yanwei Fu are with Fudan University, Shanghai, China.
E-mail: 24210980069@m.fudan.edu.cn; yanweifu@fudan.edu.cn.}
\thanks{Mengtian Li and Yuwei Lu are with Shanghai University, Shanghai, China.
E-mail: mtli@shu.edu.cn; yuilu@shu.edu.cn.}
\thanks{*Corresponding authors: Mengtian Li and Yanwei Fu.}
}

\maketitle

% ===== Teaser (forced on page 1, spans both columns, no overlap) =====
\begin{strip}
\vspace{-20mm} % optional: pull teaser upward (tune if needed)
\centering
\includegraphics[width=\textwidth]{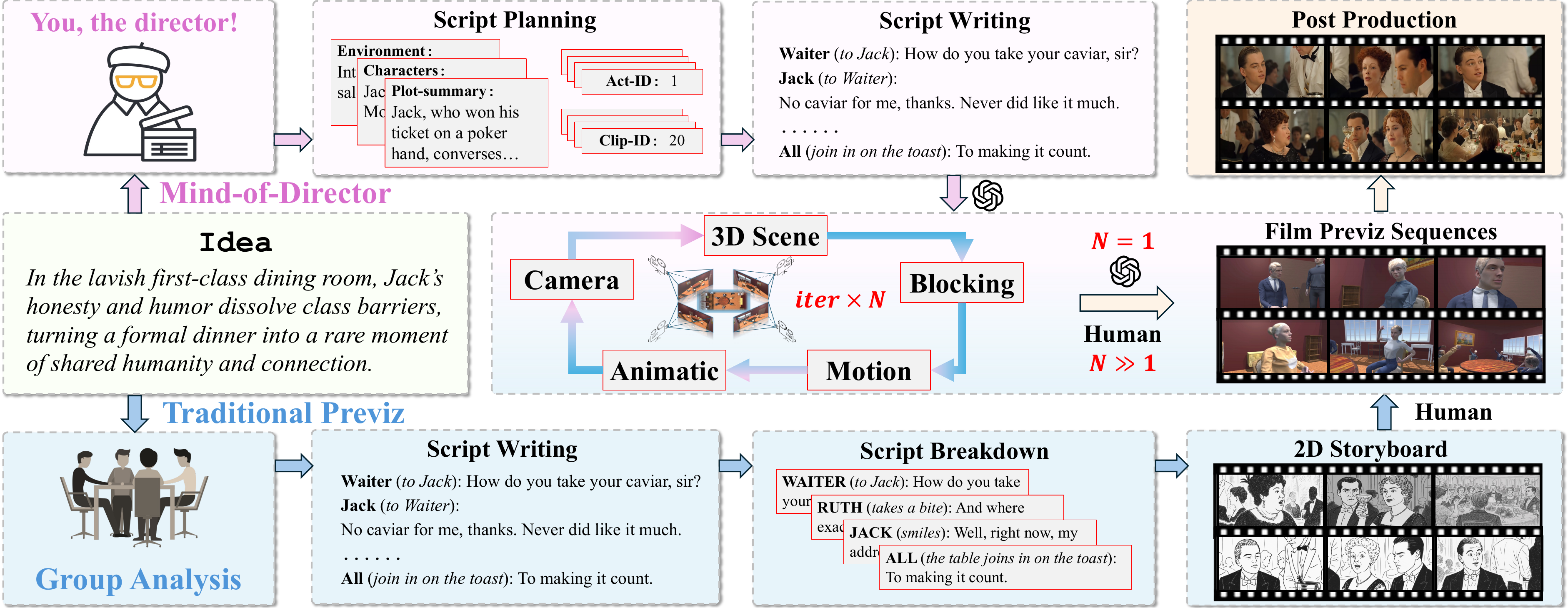}
\captionof{figure}{\textbf{Traditional Previz vs. Mind-of-Director.} %
\textit{Traditional Previz} requires iterative collaboration across multiple departments (typically iterations $N \gg 1$), involving script writing, 2D storyboarding, 3D scene construction, character blocking, animatic production, and camera planning.
In contrast, \textit{Mind-of-Director} automates this process ($N=1$) through multi-modal agents that collaborate in real-time decision-making to generate high-quality, semantically aligned, and visually coherent previz sequences directly from an idea, enabling a single creator to prototype cinematic scenes with minimal manual effort in the game engine.}
\label{fig:teaser}
\vspace{-4mm} % optional: reduce space before abstract (tune if needed)
\end{strip}

\begin{abstract}
\input{sections/00_abstract}

\end{abstract}


\IEEEpeerreviewmaketitle

% ===== Main text =====
\input{sections/01_introduction}
\input{sections/02_related_work}
\input{sections/03_method}

\input{sections/04_experiments}

\input{sections/05_conclusion}

% ===== References =====
\bibliographystyle{IEEEtran}
\bibliography{bibliography/main}

\clearpage
\input{supp}

\end{document}

%% file: sections/00_abstract.tex
We present \textbf{Mind-of-Director}, a multi-modal agent-driven framework for film previz that models the collaborative decision-making process of a film production team. 
Given a creative idea, \textbf{Mind-of-Director} orchestrates multiple specialized agents to produce previz sequences within the game engine. 
The framework consists of four cooperative modules: \textit{Script Development}, where agents draft and refine the screenplay iteratively; \textit{Virtual Scene Design}, which transforms text into semantically aligned 3D environments; \textit{Character Behaviour Control}, which determines character blocking and motion; and \textit{Camera Planning}, which optimizes framing, movement, and composition for cinematic camera effects. 
A real-time visual editing system built in the game engine further enables interactive inspection and synchronized timeline adjustment across scenes, behaviours, and cameras.
Extensive experiments and human evaluations show that \textbf{Mind-of-Director} generates high-quality, semantically grounded previz sequences in approximately 25 minutes per idea, demonstrating the effectiveness of agent collaboration for both automated prototyping and human-in-the-loop filmmaking.\\
Project Page: \href{https://pharlency.github.io/Mind-of-Director/}{https://pharlency.github.io/Mind-of-Director/}

%% file: sections/01_introduction.tex
\section{Introduction}
\label{sec:intro}

\IEEEPARstart{F}{ilm} previsualization (previz) is a crucial step bridging scriptwriting and principal photography, allowing directors to prototype spatial composition, character behaviour, camera movement, light design, and narrative pacing before production~\cite{katz1991film, mercado2013filmmaker}. 
Major films such as \textit{The Wandering Earth} and \textit{Avatar: The Way of Water} have shown how film previz systems aid planning for complex sequences involving staging, visual effects, and cinematography~\cite{li2022development, gauthier2013building}.
\textbf{However, current approaches face significant challenges:} (1) traditional previz pipelines are fragmented. 2D storyboards are drawn in Photoshop, 3D layouts built in Maya or Blender, motions captured or keyframed, and camera paths designed manually. This separation makes the process slow, labor-intensive, and hard to iterate, often taking days or weeks for a two-minute sequence.
(2) Recent text-to-video models, including \textit{Sora 2} and \textit{Veo 3.1}, create visually impressive clips but lack structured 3D representation, semantic control, and editability, making them unsuitable for the iterative, spatially grounded, and collaborative nature of previz.
In contrast, our framework streamlines this process within \textbf{about 25 minutes per idea}, offering an efficient solution for rapid creative prototyping.

As emphasized by Block~\cite{block2020visual}, film pre-production clearly reveals that at the core of film previz lies the director’s intent, which involves interpreting a script, envisioning spatial and emotional dynamics, and guiding a creative team in faithfully transforming texts into visual and editable representations~\cite{robotham2021cinematic}. 
From this perspective, two key questions arise: \textit{(1) Can a multi-modal ensemble of AI agents emulate a film crew and enable collaborative decision-making? (2) Can such agents understand the script and spatial context to achieve coherent and editable film previz sequences?}

To address these, we present \textbf{Mind-of-Director}, a multi-modal agent-driven framework that simulates the collaborative reasoning and decision-making process of a film production team under the guidance of a virtual director. 
Drawing inspiration from real filmmaking, where directors coordinate multiple departments through iterative communication and creative judgment, our system structures film previz into four modules: 
(1)\textit{ Script Development} transforms high-level ideas into structured screenplays through iterative feedback among virtual screenwriters, actors, and directors,
(2)\textit{ Virtual Scene Design} converts textual descriptions into coherent 3D environments using 2D-guided asset retrieval and rule-based spatial layout,
(3)\textit{ Character Behaviour Control} handles character blocking and motion assignment based on narrative context, 
and (4)\textit{ Camera Planning} generates cinematic shots and trajectories that express directorial intent. By maintaining a shared context and continuous feedback, these agents collaboratively produce semantically consistent, and editable film previz sequences that mirror real-world production workflows.

To support interactive creation and user supervision, the entire pipeline is fully integrated into a visual system built within Unity. 
The system provides synchronized timeline control for characters and cameras, allowing real-time adjustment of layouts, motion timing, and shot composition while maintaining narrative coherence. 
This human-in-the-loop design unifies automated generation with creative oversight, enabling efficient, iterative film previz that combines AI precision with human artistic control, completing a full sequence within about 25 minutes per idea on standard consumer hardware setups.

Our contributions are summarized as follows:
\begin{itemize}[topsep=0pt, itemsep=0em, parsep=0pt, partopsep=0pt, left=0em]

    \item We present \textbf{Mind-of-Director}, the first multi-modal agent-driven framework that transforms high-level creative ideas into editable film previz sequences through collaborative decision-making. We build \textbf{PrevizPro}, a human-annotated benchmark dataset of 360 clips with motion and camera annotations for fair evaluation of our framework.

    \item We build a multi-modal pipeline to integrate 2D-guided and rule-based strategies for efficient scene generation and optimization-based character blocking for coherent spatial arrangement. 
    We employ two collaboration mechanisms, \textit{Discuss-Revise-Judge} and \textit{Debate-Judge-Validation}, to iteratively refine the entire process, substantially improving decision quality and cross-module consistency.

    \item We develop a Unity-based system for real-time visualization, timeline editing, and human-AI co-creation, allowing a single creator to generate a full film previz sequence in approximately 25 minutes compared to the traditional 6--10 person workflow requiring 3-5 days.

\end{itemize}

%% file: sections/02_related_work.tex
{
\frenchspacing
\section{Related Work}
\label{sec:related_work}

\noindent\textbf{Storyboard \& Storytelling.}~
Storyboarding is a fundamental stage in early visual planning and cinematic concept development. 
Early methods extracted keyframes to summarize video content~\cite{bhaumik2015real, mohanta2013novel}, while recent systems adopted interactive, engine-based workflows~\cite{rao2024scriptviz, rao2023dynamic, evin2022cine} to simulate shots. 
Papalampidi et al.~\cite{papalampidi2023finding} introduced graph-based reasoning for shot selection. 
Recent advances extend storyboarding to end-to-end 2D storytelling and animation driven by LLMs~\cite{zheng2025contextualstory, yang2025seed, kim2025visagent, li2024anim, wu2025automated, wang2025mavis, shi2025animaker, he2025dreamstory, huang2025filmaster}, enabling text-to-video generation but remaining limited to planar or keyframe-based representations. 
Beyond 2D, emerging works explore 3D story representation and agent-based production frameworks~\cite{huang2024story3d, xu2024filmagent}, while commercial models such as \textit{Sora 2} and \textit{Veo 3.1} generate compelling 2D clips but still lack spatial reasoning and editability. 

Our approach advances these efforts toward fully 3D previz, employing multi-modal, multi-agent collaboration to jointly reason over scripts, spatial layouts, and cinematographic intent, producing coherent, editable film previz sequences that unify narrative and spatial design.

\vspace{3pt}
\noindent\textbf{Virtual Scene Design.}~
Early scene generation approaches relied on procedural pipelines for environment synthesis and rule-based spatial composition~\cite{deitke2022, raistrick2024infinigen}, while 3D-GPT~\cite{sun20253d} introduced LLM-based procedural modeling for language-guided generation. 
Recent training-free methods employ pretrained LLMs to synthesize 3D layouts from text~\cite{feng2023layoutgpt, yang2024holodeck, sun2025layoutvlm, huang2025fireplace}. 
In parallel, single-image-guided techniques~\cite{meng2025scenegen, bian2025holodeck, yao2025cast, ling2025scenethesis} optimize geometry and semantics for improved realism and controllability. 
StageDesigner~\cite{gan2025stagedesigner} produces stylized stage layouts from scripts. 
However, language-driven layouts often lack spatial realism, while SDF-based optimization~\cite{yao2025cast, ling2025scenethesis} achieves higher fidelity at the expense of efficiency.

To balance realism and speed, we employ an image-guided strategy that generates 2D spatial priors and constructs scene graphs with hierarchical placement rules, ensuring consistent yet efficient scene generation.

\vspace{3pt}
\noindent\textbf{Camera Planning and Cinematic Control.}~
Early studies on camera control employed rule-based frameworks for automatic cinematography and constrained shot sequencing. 
Christianson et al.~\cite{christianson1996declarative} introduced a declarative system for shot composition and transitions, inspiring later methods such as Previs-Real~\cite{qu2024previs}, which incorporated cinematographic rules into interactive previz for structured shot design. 
With the rise of large language and diffusion models, research has shifted toward language-guided and data-driven camera planning~\cite{liu2024chatcam, courant2024exceptional, chen2024cinepregen, jiang2024cinematographic}, where large models interpret text to generate trajectories. 
For evaluation, CameraBench~\cite{lin2025towards} benchmarks motion understanding, and VEU-Bench~\cite{li2025veu} measures trajectory realism and smoothness. 
However, such generative methods often produce unstable or physically inconsistent paths. 

In contrast, our method adopts a template-based strategy within a game engine, where a multi-modal model selects and parameterizes camera templates for each shot, and engine-level simulation validates trajectory safety, stability, and cinematic plausibility under physical constraints.
}

%% file: sections/03_method.tex
\section{Methodology}

\textbf{Problem Statement.}~
We formulate film previz as a collaborative generative process that transforms a high-level creative idea \(I\) into a rendered previz video \(V\). 
Formally, the process can be expressed as a composite function:
\setlength{\abovedisplayskip}{5pt}
\setlength{\belowdisplayskip}{5pt}
\begin{equation}
V = \mathcal{G}(I) = g_4(g_3(g_2(g_1(I)))),
\end{equation}
where \(\mathcal{G}\) represents our agent-driven framework that emulates the collaborative decision-making workflow of a film crew guided by a virtual director. Each sub-function \(g_i\) corresponds to a key stage in the previz pipeline as follows.

\textit{\textbf{Script Development.}}~\(S = g_1(I)\): this stage transforms a creative idea \( I \) into a structured screenplay through collaboration among the \textit{Screenwriter}, \textit{Actors}, and \textit{Director}. 
Following a \textit{discuss-revise-judge} loop, the \textit{Screenwriter} and \textit{Actors} refine the draft together, and the \textit{Director} makes the final judgment to ensure coherence, producing a screenplay \(S\) that aligns narrative structure with character intent.

\textit{\textbf{Virtual Scene Design.}}~\(E = g_2(S)\): this stage constructs a spatially coherent 3D virtual environment based on the screenplay \(S\). 
It translates narrative descriptions into a consistent and semantically rich 3D scene layout \(E\) through 2D image-guided generation, object retrieval, scene graph reasoning, and hierarchical placement rules.

\textit{\textbf{Character Behaviour Control.}}~\(B = g_3(S, E)\): this stage plans character behaviours \(B\) within the virtual environment through collaborative decision-making. 
The \textit{Director}, \textit{Cinematographer}, \textit{Scene Designer}, and \textit{Actors} first discuss blocking to ensure narrative intent, visual balance, and spatial plausibility. 
Subsequently, the \textit{Screenwriter}, \textit{Actors}, and \textit{Director} collaborate on motion selection, refining actions to align with dialogue, emotion, and story rhythm.

\textit{\textbf{Camera Planning.}}~\(V = g_4(S, E, B)\): this stage determines camera composition and trajectory through collaboration between the \textit{Director} and \textit{Cinematographer}. 
Following a \textit{debate-judge-validation} cycle, cinematographers propose and review candidate shots, while the director makes the final decision. 
Engine-level simulation then validates the selected camera parameters to remove collisions, occlusions, and instability, yielding the final previz output \(V\).

\begin{figure*}[t]
    \centering
    \includegraphics[width=\linewidth]{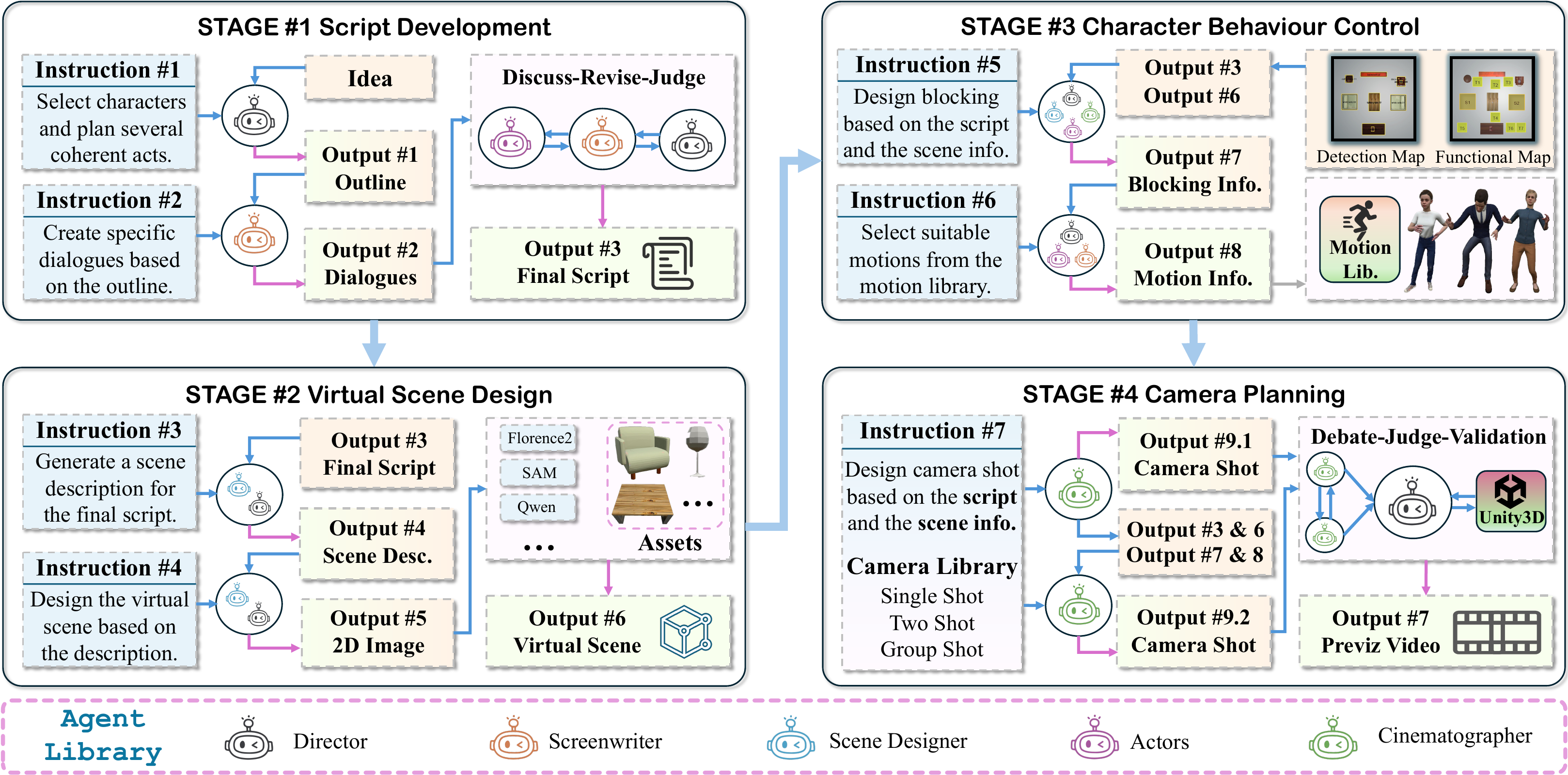}
    \caption{
    \textbf{Overview of the Mind-of-Director Framework.}~
    Given a high-level idea, our multi-modal agent-driven framework simulates a structured collaborative decision-making workflow through four interconnected modules: 
    (1) \textit{Script Development} refines the screenplay via a \textit{Discuss-Revise-Judge} process; 
    (2) \textit{Virtual Scene Design} builds consistent 3D environments using 2D-guided and rule-based generation under spatial constraints; 
    (3) \textit{Character Behaviour Control} optimizes character blocking and motion through agent feedback; 
    (4) \textit{Camera Planning} selects and validates cinematic shots via a \textit{Debate-Judge-Validation} loop for physical plausibility. 
    All modules are integrated in Unity for real-time visualization and iterative refinement.
    }
    \label{fig:method}
\end{figure*}

Through this hierarchical formulation, the framework \(\mathcal{G}\) emulates previz by combining narrative refinement, spatial design, and agent-driven decision-making as shown in Fig.~\ref{fig:method}. 
By embedding collaborative decision-making into key creative stages, \textbf{Mind-of-Director} unifies cinematic reasoning with automated multi-modal generation.

\subsection{Script Development}
\label{sec:script_writing}

Given a high-level idea \( I \), this module generates a structured screenplay \( S \) through three main reasoning and refinement stages. 
(1) The system constructs a set of character profiles \( P = \{ P_1, P_2, \dots P_m\}\), each \( P_i\) describes role’s name, age, gender, occupation, personality traits, and speaking style in detail.
(2) The idea \( I \) is decomposed into several acts \(\{A_1,\dots,A_n\}\), each \( A_i \) contains a sub-topic, participating characters, a short scene description, a story plot, and a dialogue goal.  
(3) A \textit{Discuss-Revise-Judge} loop is applied to iteratively refine the screenplay, as described in Algorithm~\ref{alg:script_writing_djr}.

\begin{algorithm}[h]
\footnotesize
\caption{Discuss-Revise-Judge Process}
\label{alg:script_writing_djr}
\begin{algorithmic}[1]
\State \textbf{Notation:}
\Statex \(\;I\): high-level idea \quad \(S\): screenplay \quad \(A_i\): the \(i\)-th act
\Statex \(\;W\): Screenwriter \quad \(A\): Actors \quad \(D\): Director
\Statex \(\;D_i^{(t)}\): draft of act \(A_i\) at iteration \(t\) \quad \(F_i^{\text{actor}}\), \(F_i^{\text{dir}}\): feedback
\Statex \(\;\Phi(\cdot,\cdot)\): revision operator
\vspace{4pt}

\For{each act \(A_i\) derived from \(I\)}
    \State \(W: I \xrightarrow{} D_i^{(0)}\) \hfill \(\triangleright\) initial draft
    \Repeat
        \State \(D_i^{(t)} \xrightarrow{A} F_i^{\text{actor}}\)  \hfill \(\triangleright\) actors discuss and feedback
        \State \(D_i^{(t)}, F_i^{\text{actor}} \xrightarrow{W:\,\Phi} D_i^{(t+1)}\) \hfill \(\triangleright\) screenwriter revision
        \State \(D_i^{(t+1)} \xrightarrow{D} F_i^{\text{dir}}\) \hfill \(\triangleright\) director judgment
        \If{\(D\) approves}
            \State \(D_i^{*} \leftarrow D_i^{(t+1)}\)
        \Else
            \State \(D_i^{(t+1)}, F_i^{\text{dir}} \xrightarrow{W:\,\Phi} D_i^{(t+2)}\)
        \EndIf
    \Until{approval}
    \State \(S \leftarrow S \cup \{(A_i, D_i^{*})\}\)
\EndFor
\State \textbf{Output:} final screenplay \(S\)
\end{algorithmic}
\end{algorithm}

\subsection{Virtual Scene Design}
\label{sec:virtual_scene_design}

\noindent\textbf{Challenges.}~
Existing scene generation methods face key limitations. 
LLM-based layouts often produce unrealistic placements and inconsistent spatial relations due to the absence of visual priors. 
Optimization-based approaches such as SDF-driven reconstruction~\cite{ling2025scenethesis, yao2025cast} improves spatial accuracy but is computationally heavy and impractical for real-time previz. 
End-to-end pipelines~\cite{liang2024luciddreamer, wu2024reconfusion} such as Gaussian splatting, yield photorealistic but non-editable assets that lack the geometric consistency for production.

To address these issues, we adopt StageDesigner~\cite{gan2025stagedesigner} as the backbone architecture and further incorporate explicit visual priors into the pipeline, enabling more realistic and controllable scene generation while preserving structural efficiency.

\vspace{1pt}
\noindent\textbf{Guidance Image Generation.}~
First, generate a 2D guidance image \(G_i\) from the scene description \(SD_i\) of each act \(A_i\). 
The image provides a spatial prior encoding composition and object arrangement, serving as the foundation for subsequent object retrieval and spatial layout reasoning.

\vspace{1pt}
\noindent\textbf{Object Retrieval.}~
We retrieve high-quality, production-ready assets with editable meshes, UV maps, and PBR materials from the curated Objaverse~\cite{deitke2023objaverse} subset refined by Holodeck~\cite{yang2024holodeck}. 
Given \(G_i\), objects are detected using Florence2~\cite{xiao2024florence} and segmented by SAM~\cite{kirillov2023segment}. 
Since detection may yield invalid, fragmented, or occluded instances, qwen3-max filters redundant segments, and qwen-image-edit inpaints missing regions to restore object integrity and visual coherence. 
For each completed object, qwen3-max generates detailed textual descriptions, and multi-modal similarity is computed by combining CLIP image–text and Sentence-BERT text–text scores. 
The asset with the highest combined score is selected, ensuring semantic alignment and efficient retrieval.

\vspace{1pt}
\noindent\textbf{Base Environment Construction.}~
Retrieved assets are organized using a scene graph extracted from the detection image, encoding: 
(1) classification into anchor and non-anchor objects, 
(2) geometric attributes such as dimensions, orientation and scale,
(3) spatial relations of adjacency, support, and relative positioning. 
To reduce uncertainty in VLM inference, we apply an ensemble strategy that aggregates multiple reasoning trials and retains consistent relations, yielding a stable scene graph. 
Based on this graph, we construct a hierarchical occupancy grid map for efficient asset placement. 
A ground-level \(M \times N\) grid (scene floor) defines anchor placements, while each anchor’s top surface generates a secondary grid for non-anchor objects. 
This layered formulation enables efficient, collision-free asset arrangement without iterative optimization and provides spatial priors for subsequent behaviour control.

\vspace{1pt}
\noindent\textbf{Ornament Enhancement.}~
As single-view images lack global context, we use LLM reasoning to populate decorative and background elements (e.g., plants, lamps, books) into unoccupied yet contextually suitable grid regions. 
This step enriches composition, enhances visual balance, and improves realism, yielding the final 3D environment \(E_i\).

\subsection{Character Behaviour Control}
\label{sec:behaviour_control}
\noindent\textbf{Challenges.}~
Generating realistic and meaningful character behaviours in 3D space remains difficult. 
Directly predicting performing positions with LLMs is unreliable, often resulting in physically or cinematically invalid placements. 
FilmAgent~\cite{xu2024filmagent} depends on manually predefined candidate points, disrupting automation and limiting scalability. 

We therefore adopt a hybrid pipeline that first optimizes valid performing regions within the scene, then uses a multi-modal LLM to select the most appropriate positions guided by spatial and narrative cues as follows.

\noindent\textbf{Behaviour Modeling.}~
First, the behaviour of each character in the scene can be represented formally as
\setlength{\abovedisplayskip}{4pt}
\setlength{\belowdisplayskip}{4pt}
\begin{equation}
B_{ijk} =
  (S^{\mathrm{start}}_{ijk}, P^{\mathrm{start}}_{ijk}, F^{\mathrm{start}}_{ijk},
   S^{\mathrm{end}}_{ijk}, P^{\mathrm{end}}_{ijk}, F^{\mathrm{end}}_{ijk}),
\end{equation}
where \(S\), \(P\), and \(F\) denote the character’s physical state (standing or sitting), spatial position, and facing direction at both the beginning and end of each dialogue-driven clip.

\vspace{1pt}
\noindent\textbf{Performing Region Optimization.}~
We extend the occupancy grid map from Sec.~\ref{sec:virtual_scene_design} to identify valid standing regions.  
Each candidate region \(R_i\), defined as a \(60\,\text{cm}\!\times\!60\,\text{cm}\) obstacle-free square, is assigned a scalar loss:
\setlength{\abovedisplayskip}{4pt}
\setlength{\belowdisplayskip}{4pt}
\begin{equation}
L_i = L_{\text{coll}}(c_i) + L_{\text{cam}}(c_i),
\label{eq:Loss}
\end{equation}
where \(L_{\text{coll}}\) penalizes boundary violations or object collisions, and \(L_{\text{cam}}\) rewards camera visibility.  
Specifically,
\setlength{\abovedisplayskip}{4pt}
\setlength{\belowdisplayskip}{4pt}
\begin{align}
L_{\text{coll}}(c_i) &=
w_b e^{-d_B(c_i)/\sigma_B}
+ w_o e^{-d_O(c_i)/\sigma_O}, \label{eq:Lcoll}\\
L_{\text{cam}}(c_i) &=
w_c \bigl(1-\min(\bar S(c_i), S_{\max})\bigr)^{\alpha_{\text{cam}}}, \label{eq:Lcam}
\end{align}
where \(d_B\) and \(d_O\) denote the distances to the nearest boundary and obstacle,  
\(\bar S(c_i)=\tfrac{1}{4}\!\sum_{k=1}^{4}S_k(c_i)\) is the average visibility ratio from four cameras,  
\(w_b,w_o,w_c\) are weights,  
\(\sigma_B,\sigma_O\) are distance scales,  
\(S_{\max}\) the visibility cap,  
and \(\alpha_{\text{cam}}\) the visibility penalty exponent.  
To ensure spatial safety and cinematic readability, a threshold \(\tau\) is defined, and within each \(1\,\text{m}\times1\,\text{m}\) region \(P_j\),  
we select the optimal performing zone with the minimum loss satisfying \(L_i<\tau\):
\setlength{\abovedisplayskip}{4pt}
\setlength{\belowdisplayskip}{4pt}
\begin{equation}
R_j^*=\arg\min_{R_i\subseteq P_j,\,L_i<\tau} L_i.
\end{equation}
For sitting, the LLM directly assigns a sittable attribute to items by reasoning over object names in the scene graph.

\vspace{1pt}
\noindent\textbf{Performing Region Selection.}~
The optimized standing and inferred sitting regions are projected into two top-view maps:  
a scene detection map \(DM_i\) with object bounding boxes and semantic labels, and a functional scene map \(FM_i\) marking valid performing regions under spatial constraints.  
Together with dialogue content and cinematic priors, these are fed into a multi-modal LLM to infer each character’s behaviour \(B_{ijk}\).  
The model combines visual and textual reasoning to select positions maximizing narrative relevance and cinematic clarity, which are further refined through multi-agent collaboration among the \textit{Director}, \textit{Cinematographer}, \textit{Scene Designer}, and \textit{Actors} (details in the supplementary material).

\vspace{1pt}
\noindent\textbf{Motion Selection.}
After spatial blocking is finalized, each character is assigned an appropriate motion from a curated Mixamo animation library based on dialogue context and behaviour \(B_{ijk}\). 
The selected animations are retargeted to the predefined character rigs to ensure structural compatibility and consistent skeletal articulation. 
The \textit{Screenwriter}, \textit{Actors}, and \textit{Director} collaboratively refine these motion choices through a \textit{Discuss-Revise-Judge} loop to maintain narrative coherence and expressive consistency. 
For scene-level locomotion and navigation, we employ Unity’s built-in NavMesh system to compute collision-aware and physically feasible paths within the 3D environment under dynamic spatial constraints. 
This ensures that character movements respect scene geometry, obstacle constraints, and occlusion boundaries, resulting in spatially grounded and executable motion sequences.

\subsection{Camera Planning}
\label{sec:camera_planning}

\noindent\textbf{Challenges.}~
Existing camera planning methods still face two key limitations in practical cinematic deployment. 
Generative approaches relying solely on language or diffusion models~\cite{liu2024chatcam, courant2024exceptional, chen2024cinepregen, jiang2024cinematographic} often produce unstable or physically inconsistent trajectories. 
FilmAgent~\cite{xu2024filmagent} uses predefined camera positions, but this restricts flexibility and overlooks the game engine’s built-in physics for collision and occlusion detection.

We therefore construct a camera template library, employ a multi-modal LLM to select the most suitable shots, and adopt the \textit{Debate-Judge-Validation} process for iterative shot refinement. Details are as follows.

\vspace{1pt}
\noindent\textbf{Camera Template Library.}~
Following the cinematic principles discussed in Giannetti’s \textit{Understanding Movies}~\cite{giannetti2005understanding}, 
we construct a hierarchical, parameterized camera library comprising 12 single-person, 8 two-person, and 1 group shot categories at a high level. 
Rather than fixed camera presets, each template is defined as a Python-style function \(T_k(\Theta_k)\), where \(\Theta_k\) specifies a set of controllable parameters (e.g., viewpoint, distance, and angle) with usage rules and applicability tags (details in the supplementary material).

\vspace{1pt}
\noindent\textbf{Shot Selection.}~
For each clip \(C_{ij}\), two top-view maps from Sec.~\ref{sec:behaviour_control} (detection and functional map), along with blocking information, screenplay context, and the camera template library, are input to a multi-modal LLM to infer the suitable camera template \(T_k\) and parameters \(\Theta_k\).  
To emulate collaborative cinematographic decision-making, we adopt a \textit{Debate-Judge-Validation} strategy (Algorithm~\ref{alg:debate_judge_validation}). 
Two \textit{cinematographers} independently propose templates, critique each other’s designs, and provide feedback. 
The \textit{Director} aggregates the results, makes the final selection, and submits it to the engine for physical validation. 
If collisions, occlusions, or framing issues are detected, the engine returns feedback for parameter refinement until a valid configuration is achieved.

\begin{algorithm}[t]
\footnotesize
\caption{Debate-Judge-Validation Process}
\label{alg:debate_judge_validation}
\begin{algorithmic}[1]
\State \textbf{Notation:} 
\Statex \(\;P\): Cinematographers \quad \(D\): Director
\Statex \(\;DM_i\): Scene detection map \quad \(FM_i\): Functional scene map
\Statex \(\;B_{ij}\): blocking information \quad \(S\): Screenplay
\Statex \(\;CT\): Camera library \quad \(T_k(\Theta_k)\): Camera with parameters
\vspace{4pt}

\For{each clip \(C_{ij}\) in act \(A_i\)}
    \State \(\{DM_i, FM_i, B_{ij}, S, CT\} \xrightarrow{\text{encode}} \text{MLLM}\)
    \State \(\text{MLLM} \xrightarrow{} \{T_k(\Theta_k)\}\)
    \State \(P_1, P_2 \xrightarrow{\text{independent proposal}} T^{(1)}_{ij}, T^{(2)}_{ij}\)
    \State \(P_1 \leftrightarrow P_2 \xrightarrow{\text{debate}} F^{(1)}_{ij}, F^{(2)}_{ij}\)
    \State \(\{T^{(1)}_{ij},T^{(2)}_{ij},F^{(1)}_{ij},F^{(2)}_{ij}\} \xrightarrow{D} (T^*_{ij}, \Theta^*_{ij})\)
    \State \((T^*_{ij}, \Theta^*_{ij}) \xrightarrow{\text{engine validation}} \text{collisions, occlusions}\)
    \If{validation passes}
        \State \(T^*_{ij} \Rightarrow \text{final shot configuration}\)
    \Else
        \State \(\text{engine feedback} \xrightarrow{} D \xrightarrow{\text{adjust}} (T', \Theta')\)
        \State Repeat validation until convergence
    \EndIf
\EndFor
\State \(\{(T^*_{ij}, \Theta^*_{ij})\}_{i,j} \Rightarrow \text{final camera plan}\)
\end{algorithmic}
\end{algorithm}

% \begin{tcolorbox}[title=\textbf{Example Camera Template: TWOSTATIC},
% colback=white, colframe=black!30, arc=2mm, boxrule=0.4pt, breakable]
% \scriptsize
% \begin{verbatim}
% def TWOSTATIC(A, B, relation, framing, size, angle):
%     """
%     Intent: Present relationship balance or tension.
%     Description: Two subjects framed statically.
%     Use When: Dialogue, reconciliation, argument.
%     Output:
%     {
%       "type": "two_static",
%       "subjects": [A, B],
%       "relation": relation,
%       "framing": framing,
%       "shot_size": size,
%       "angle": angle,
%       "rationale": "Static composition 
%                     defines relational geometry."
%     }
%     RELATIONAL EXTENSIONS:
%       relation: {"equal","dominant_A","dominant_B",
%                  "intimate","distant","conflict"}
%       framing: {"two_shot","OTS_pair","profile_duet",
%                 "split_depth","cross_cut"}
%       focus_mode: {"both","focus_A","focus_B",
%                    "rack_between"}
%     """
% \end{verbatim}
% \end{tcolorbox}

%% file: sections/04_experiments.tex
\section{Experiments}
\label{sec:experiments}

\subsection{Experiment Setup}
\label{sec:experiment_setup}

\noindent\textbf{Dataset.}~
We build \textbf{PrevizPro}, a human-annotated dataset for evaluating our framework. 
It contains ten creative ideas expanded into three acts each, resulting in 30 acts and 360 clips. 
For every clip, experts annotate character motions and assign suitable camera templates matching the narrative context. 
As our framework is training-free, PrevizPro serves solely as an evaluation benchmark for all modules.

\vspace{1pt}
\noindent\textbf{Evaluation metrics.}~
Following real-world previz workflows, we evaluate four stages: \textit{Script Development}, \textit{Virtual Scene Design}, \textit{Character Behaviour Control}, and \textit{Camera Planning}, using both quantitative and human metrics.

\begin{itemize}[topsep=0pt, itemsep=0.2em, parsep=0pt, partopsep=0pt, left=0em]
\item \textbf{Script Development.}~
For human evaluation, \textit{Character Authenticity} (CA) checks dialogue–character consistency and emotional realism, \textit{Pacing} (PA) measures narrative rhythm and scene flow, and \textit{Overall Quality} (OA) assesses script coherence and overall cinematic readability.

\item \textbf{Virtual Scene Design.}~
Quantitatively, CLIP~\cite{radford2021learning} evaluate text–scene alignment, 
\textit{Collision Rate} (Col.) measures the percentage of object collisions, 
and \textit{Reach}/\textit{Walk} from Physcene~\cite{yang2024physcene} assess interactivity.
For human evaluation, we use \textit{Visual Quality} (VQ), \textit{Physical Plausibility} (PP), and \textit{Script Alignment} (SA).

\item \textbf{Behaviour Control.}~
For blocking, Loss in Equation (\ref{eq:Loss}) quantifies spatial quality, normalized to \([0,1]\) by \(\frac{L-L_{\min}}{L_{\max}-L_{\min}}\). 
For human evaluations, \textit{Positional Plausibility} (Pos-P) assesses how naturally characters are placed.  
For motion, \textit{Accuracy} (Acc.) measures the ratio of correctly predicted motion IDs, while \textit{Diversity} (\(\text{Div.}=\frac{-\sum_i p(a_i)\log_2 p(a_i)}{\log_2 n}\)) captures expressive variety via normalized Shannon entropy,  
where \(p(a_i)\) is the frequency of motion \(a_i\) and \(n\) is the number of distinct motions.

\item \textbf{Camera Planning.}~
Quantitatively, \textit{Collision Rate} (Col.) measures the proportion of camera trajectories intersecting scene objects, and \textit{Occlusion Rate} (Occ.) quantifies frames where key characters are obscured. 
For human evaluation, \textit{Accuracy} (Acc.) compares predicted camera selections with human-annotated templates, while \textit{Overall Quality} (OQ) assesses cinematic appeal, composition, and motion smoothness. 
Each clip is annotated with two optimal camera templates serving as references.
\end{itemize}

\vspace{1pt}
\noindent\textbf{Baselines.}~
As most existing systems focus on isolated aspects of AI-driven film previz rather than a full collaborative pipeline, we compare our framework with two representative methods: \textit{StageDesigner}~\cite{gan2025stagedesigner} and \textit{FilmAgent}~\cite{xu2024filmagent}. 
StageDesigner emphasizes script-to-3D scenographic layout through language reasoning, while FilmAgent generates character behaviour and camera planning purely through LLM-driven reasoning. 
In our implementation, we follow FilmAgent’s original language-based formulation and provide scene objects as top-view bounding box coordinates serialized into textual descriptions, without introducing additional visual priors. 
To ensure a fair comparison, we adapt the StageDesigner backbone to our task setting while preserving its original LLM-driven layout generation mechanism without incorporating visual priors. 
This allows us to isolate the effect of visual prior integration and directly evaluate its contribution to spatial realism and controllability. 
These baselines highlight the contrast between text-only reasoning pipelines and our multi-modal framework, which integrates spatial reasoning, narrative coherence, and cinematic intent for a controlled comparison between unimodal and multimodal previz.

\vspace{1pt}
\noindent\textbf{Configurations.}~
Qwen3-max serves as the core foundation model across all modules, with qwen-image-plus for image generation and qwen-image-edit for detailed inpainting during virtual scene design. 
CosyVoice synthesizes synchronized dialogue audio for all characters. 
The entire framework operates in a training-free manner, with agent interactions conducted for up to 3 iterative rounds per stage.
For the experimental table, Note that ``\textasteriskcentered{}'' indicates results obtained from our experiments. For all metrics, $\uparrow$/$\downarrow$ denotes higher/lower is better, and human evaluation scores are reported as ``mean (standard deviation)''.

\begin{figure*}[t]
    \centering
    \includegraphics[width=\linewidth]{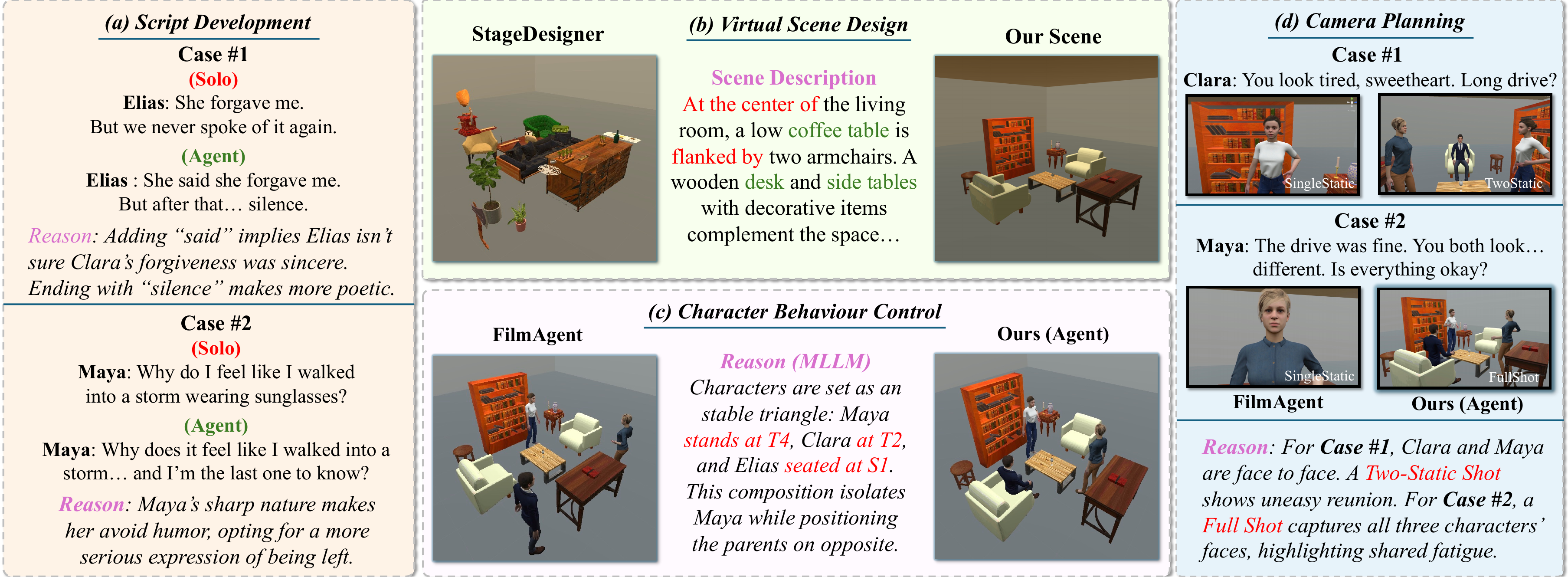}
    \caption{
    \textbf{Qualitative Comparison.}~
    We present a representative sample from Act \(A_i\) to demonstrate our framework's performance and cross-stage consistency. The image shows results across four stages: 
    (a) \textit{Script Development}: Comparison of screenplay generated by Solo vs. Agent Collaboration; 
    (b) \textit{Virtual Scene Design}: Comparison of scene layouts from StageDesigner and our approach with improved spatial grounding; 
    (c) \textit{Character Behaviour Control}: Character positioning from FilmAgent vs. our agent-driven method; 
    (d) \textit{Camera Planning}: Camera shot selection, comparing FilmAgent and our approach.
    }
    \label{fig:pic}
\end{figure*}

\subsection{Quantitative Analysis}

\noindent\textbf{Virtual Scene Design.}~
As shown in Table~\ref{tab:virtualscenedesign}, our method outperforms the baseline in film previz, achieving stronger text–scene alignment (CLIP: \textbf{26.35} vs.\ 25.48) and reducing collision rate from 2.27\% to \textbf{0.83\%}. It also improves scene interactivity, with higher Reach (\textbf{76.32\%} vs.\ 53.58\%) and Walk (\textbf{94.32\%} vs.\ 72.65\%), indicating better accessibility and navigability in 3D environments for previz tasks. 

During experimentation, we observe that purely LLM-driven layout generation tends to increase object density within scenes, as language models often introduce additional assets to enrich visual descriptions. While this enhances apparent visual richness, it frequently leads to cluttered spatial arrangements that degrade functional accessibility, thereby reducing Reach and Walk metrics. In contrast, integrating visual priors provides grounded spatial constraints and object co-occurrence guidance, encouraging more balanced object selection and placement. As a result, the generated scenes achieve a better trade-off between visual expressiveness and functional usability, improving both realism and navigability.

\begin{table}[t]
\centering
\footnotesize
% \setlength{\tabcolsep}{3pt}        % tighter columns
% \renewcommand{\arraystretch}{1.05} % tighter rows

% ---------- (a) Script Development ----------
\captionof{table}{Results on \textbf{Script Development}. Experts and users conduct a comparative evaluation under both Solo and Agent settings. 
\textit{Solo} corresponds to a single screenwriter baseline, where the screenplay is generated in one forward LLM pass without multi-agent reasoning or iterative revision.}
\label{tab:scriptdevelopment}
\vspace{-3pt}
\begin{tabularx}{\linewidth}{
>{\centering\arraybackslash}p{0.16\linewidth}
>{\centering\arraybackslash}p{0.16\linewidth}
>{\centering\arraybackslash}p{0.15\linewidth}
>{\centering\arraybackslash}p{0.15\linewidth}
>{\centering\arraybackslash}p{0.15\linewidth}
}
\toprule
\textbf{Evaluator} & \textbf{Method} & \textbf{CA} & \textbf{PA} & \textbf{OA} \\
\midrule
\multirow{2}{*}{Expert} 
    & Solo    & 29.2\% & 15.8\% & 19.2\% \\
    & Agent   & \textbf{70.8\%} & \textbf{84.2\%} & \textbf{80.8\%} \\
\midrule
\multirow{2}{*}{User}   
    & Solo    & 42.8\% & 23.0\% & 24.8\% \\
    & Agent   & \textbf{57.2\%} & \textbf{77.0\%} & \textbf{75.2\%} \\
\bottomrule
\end{tabularx}

\vspace{6pt}

% ---------- (b) Virtual Scene Design ----------
\captionof{table}{Results on \textbf{Virtual Scene Design}. Quantitative metrics evaluate semantic similarity, spatial plausibility, and scene interactivity. Human evaluation assesses \textit{Visual Quality} (VQ), \textit{Physical Plausibility} (PP), and \textit{Script Alignment} (SA).}
\label{tab:virtualscenedesign}
\vspace{3pt}

\begin{tabularx}{\linewidth}{
>{\centering\arraybackslash}p{0.22\linewidth}
>{\centering\arraybackslash}p{0.14\linewidth}
>{\centering\arraybackslash}p{0.14\linewidth}
>{\centering\arraybackslash}p{0.14\linewidth}
>{\centering\arraybackslash}p{0.14\linewidth}
}
\toprule
\textbf{Method} & \textbf{CLIP}$\uparrow$ & \textbf{Col.}$\downarrow$ & \textbf{Reach}$\uparrow$ & \textbf{Walk}$\uparrow$ \\
\midrule
StageDesigner* & 25.48 &  2.27\% & 53.58\% & 72.65\% \\
Ours           & \textbf{26.35} & \textbf{0.83\%} & \textbf{76.32\%} & \textbf{94.32\%} \\
\bottomrule
\end{tabularx}

\vspace{4pt}

\begin{tabularx}{\linewidth}{
>{\centering\arraybackslash}p{0.15\linewidth}
>{\centering\arraybackslash}p{0.18\linewidth}
>{\centering\arraybackslash}p{0.15\linewidth}
>{\centering\arraybackslash}p{0.15\linewidth}
>{\centering\arraybackslash}p{0.15\linewidth}
}
\toprule
\textbf{Evaluator} & \textbf{Method} & \textbf{VQ} & \textbf{PP} & \textbf{SA} \\
\midrule
\multirow{2}{*}{Expert} 
    & StageDesigner* & 40.8\% & 11.7\% & 25.8\% \\
    & Ours           & \textbf{59.2\%} & \textbf{88.3\%} & \textbf{74.2\%} \\
\midrule
\multirow{2}{*}{User}   
    & StageDesigner* & 46.0\% & 16.3\% & 22.3\% \\
    & Ours           & \textbf{54.0\%} & \textbf{83.7\%} & \textbf{77.7\%} \\
\bottomrule
\end{tabularx}

\vspace{6pt}

% ---------- (c) Behaviour Control ----------
\captionof{table}{Results on \textbf{Behaviour Control}. Quantitative metrics are assessed by spatial accuracy, motion accuracy and diversity, while human evaluations measure \textit{positional plausibility} (Pos-P). 
\textit{Solo} corresponds to a single forward pass of the MLLM without the Discuss-Revise-Judge loop.}
\label{tab:characterbehaviourcontrol}
\vspace{3pt}
\begin{tabularx}{\linewidth}{
>{\centering\arraybackslash}p{0.25\linewidth}
>{\centering\arraybackslash}p{0.12\linewidth}
>{\centering\arraybackslash}p{0.16\linewidth}
>{\centering\arraybackslash}p{0.12\linewidth}
>{\centering\arraybackslash}p{0.12\linewidth}
}
\toprule
\multirow{2}{*}{\textbf{Method}} 
& \multicolumn{2}{c}{\textbf{Blocking}} 
& \multicolumn{2}{c}{\textbf{Motion}} \\
\cmidrule(lr){2-3} \cmidrule(lr){4-5}
& \textbf{Loss}$\downarrow$ & \textbf{Pos-P} & \textbf{Acc.}$\uparrow$ & \textbf{Div.} \\
\midrule
FilmAgent*    & 0.86 & 2.25 (0.97) & --      & --   \\
Ours (Solo)   & 0.52 & 3.92 (0.79) & 83.24\% & 0.65 \\
Ours (Agent)  & \textbf{0.48} & \textbf{4.08} (0.67) & \textbf{88.79\%} & \textbf{0.73} \\
\bottomrule
\end{tabularx}

\vspace{6pt}

% ---------- (d) Camera Planning ----------
\captionof{table}{Results on \textbf{Camera Planning}. Quantitative metrics are assessed by physical validity, while human evaluations focus on cinematic quality, composition, and overall visual smoothness. \textit{Solo} corresponds to a single forward pass of the MLLM without the Debate-Judge-Validation loop.}
\label{tab:cameraplanning}
\vspace{3pt}
\begin{tabularx}{\linewidth}{
>{\centering\arraybackslash}p{0.25\linewidth}
>{\centering\arraybackslash}p{0.11\linewidth}
>{\centering\arraybackslash}p{0.11\linewidth}
>{\centering\arraybackslash}p{0.11\linewidth}
>{\centering\arraybackslash}p{0.16\linewidth}
}
\toprule
\textbf{Method} & \textbf{Col.}$\downarrow$ & \textbf{Occ.}$\downarrow$ & \textbf{Acc.}$\uparrow$ & \textbf{OQ}$\uparrow$ \\
\midrule
FilmAgent*    & 13.7\% & 11.4\% & 58.3\% & 1.88 (0.83) \\
Ours (Solo)   & 9.6\%  & 7.3\%  & 64.4\% & 3.62 (0.74) \\
Ours (Agent)  & \textbf{2.1\%} & \textbf{1.6\%} & \textbf{79.2\%} & \textbf{4.12} (0.64) \\
\bottomrule
\end{tabularx}

% Optional: pull up a bit if needed
\vspace{-6pt}
\end{table}

\noindent\textbf{Character Behaviour Control.}~
As shown in Table~\ref{tab:characterbehaviourcontrol}, for blocking, our Agent framework achieves lower Loss (\textbf{0.48} vs.\ 0.86) than FilmAgent, proving that visual cues guide more suitable and cinematic character positioning to avoid collisions, boundaries, and poor visibility. For motion, our framework attains \textbf{88.79\%} accuracy and \textbf{0.73} diversity, showing its ability to capture narrative intent, assign context-appropriate actions, and maintain expressive variety.

\noindent\textbf{Camera Planning.}~
As shown in Table~\ref{tab:cameraplanning}, our method significantly outperforms the purely LLM-based FilmAgent, reducing camera collision (13.7\% vs. \textbf{2.1\%}) and occlusion (11.4\% vs. \textbf{1.6\%}) rates. These gains stem from integrating visual cues within the multi-modal LLM and leveraging the robust physics-based collision and occlusion detection of game engine. Additionally, our approach attains higher camera accuracy (\textbf{79.2\%} vs.\ 58.3\%), especially in multi-person shots, effectively capturing character positioning and facing relationships based on visual cues to ensure alignment between script intent and spatial dynamics.

\subsection{Qualitative Analysis}

We sample an Act \(A_i\) to illustrate our framework’s full workflow, as shown in Figure~\ref{fig:pic}. 
For Script Development, compared with the Solo setting, the \textit{Discuss-Revise-Judge} interaction among agents notably enhances screenplay expressiveness, aligning dialogue with character traits and narrative flow. 
In Virtual Scene Design, we compare StageDesigner and our method. While StageDesigner generates visually rich scenes, its layouts exhibit object collisions and limited walkable areas. Our 2D-guided approach extracts visual priors to produce spatially coherent, interaction-friendly environments optimized for character performance. 
For Character Behaviour Control, our method achieves more stable and realistic blocking than the LLM-driven FilmAgent, maintaining triangular formations and optimal camera angles. Lacking visual grounding, FilmAgent often misinterprets spatial relations, leading to unstable placements. 
In Camera Planning, visual cues enable our approach to capture stable formations and facing relationships between characters, effectively employing two-person or multi-person shots without predefined camera positions, yielding more flexible and cinematic compositions.

\subsection{Human Evaluation}

\noindent\textbf{Criteria and Setup.}~
We conducted a comparative user study focused on Virtual Scene Design, recruiting 12 film-school experts (6 male and 6 female, including 6 actors, 4 cinematographers, and 2 directors) and 40 general users (20 male and 20 female, with no filmmaking experience). 
We sampled ten scenes from scripts and generated scenes using both StageDesigner and our method. 
Experts and users evaluated the scenes based on \textit{Visual Quality} (VQ), \textit{Physical Plausibility} (PP), and \textit{Script Alignment} (SA).

\noindent\textbf{Evaluation Results.}~
As shown in Table~\ref{tab:virtualscenedesign}, our method received slightly higher support for Visual Quality, but demonstrated a more significant advantage in Physical Plausibility and Script Alignment. 
Experts and users favored our method for producing more spatially coherent layouts and ensuring better consistency with the script, proving that 2D visual priors effectively guide the large model to build more reasonable scenes, making it better suited for previz tasks.

\subsection{Ablation Studies}

\noindent\textbf{Quantitative Analysis.}~
For Virtual Scene Design, we have previously analyzed in Table~\ref{tab:virtualscenedesign} the effect of incorporating visual priors into the StageDesigner backbone. 
This comparison effectively serves as an ablation study isolating the contribution of visual priors. For Character Behaviour Control, in blocking, the agent method outperforms the solo approach, reducing the loss from 0.52 to \textbf{0.48}, highlighting the improvement in position selection through the \textit{Discuss-Revise-Judge} interaction. 
In motion, accuracy increased from 83.24\% (solo) to \textbf{88.79\%} (agent), and diversity from 0.65 to \textbf{0.73}. 
For \textbf{Camera Planning}, the agent-based \textit{Debate-Judge-Validation} interaction reduced collision rate from 9.6\% (solo) to \textbf{2.1\%}, occlusion from 7.3\% to \textbf{1.6\%}, and improved accuracy from 64.4\% to \textbf{79.2\%}. 
These results demonstrate that collaborative decision-making and engine-level constraints substantially enhance spatial validity and cinematic quality.

\noindent\textbf{Human Evaluation.}~
For Script Development, we sampled ten scripts and asked experts and users to evaluate them in both solo and agent-based settings. As shown in Table~\ref{tab:scriptdevelopment}, both experts and users favored the agent-driven scripts, with improvements in all metrics.
For Character Blocking, experts rated the agent-driven method higher, increasing \textit{Positional Plausibility} from 3.92 (solo) to \textbf{4.08}, indicating better alignment with narrative intent (Table~\ref{tab:characterbehaviourcontrol}).
For Camera Planning, the agent’s \textit{debate-judge-validation} process outperformed the solo method, scoring \textbf{4.12} vs. 3.62 as shown in Table~\ref{tab:cameraplanning}, suggesting that agent collaboration results in a more cinematic and smooth composition.

%% file: sections/05_conclusion.tex
\section{Discussion}

The consistent improvements observed across modules suggest the effectiveness of structured multi-agent collaboration for complex cinematic planning tasks. Rather than treating large models as isolated generators, our framework introduces explicit role specialization and iterative evaluation, which collectively reduce error propagation across stages and improve constraint satisfaction in spatial and narrative reasoning under multi-stage dependencies.
Beyond quantitative gains, the results indicate that collaborative reasoning provides a principled mechanism for decomposing creative tasks into structured sub-decisions. By distributing responsibilities across agents and enforcing revision–validation loops, the system maintains coherence across script, layout, behaviour, and camera planning. This structured interaction offers a computational analogue to distributed creative workflows, providing insights into how generative models can be organized to support higher-level design objectives.
The framework also demonstrates potential for adaptation to related domains such as interactive storytelling, game prototyping, and immersive AR/VR scene design. However, such extensions require careful consideration of scalability, asset diversity, and domain-specific constraints.

\section{Limitations and Future Work}

While the framework performs reliably in moderately complex scenarios, scalability remains an open challenge. As scene size, object count, and character interactions increase, the combinatorial reasoning space of multi-agent collaboration expands significantly, leading to higher inference latency and computational cost. In particularly dense layouts or tightly constrained environments, the system may produce suboptimal arrangements that require manual refinement.
Although spatial feasibility is enforced through engine-level tools such as NavMesh and collision detection, the framework does not yet incorporate fine-grained physical simulation or differentiable constraint optimization. Consequently, detailed contact dynamics and complex human–object interactions remain approximated rather than physically modeled.
Character performance is currently bounded by curated animation libraries and retargeting pipelines. While stable, this design limits expressive diversity and stylistic nuance. Similarly, camera planning relies on structured template families; although parameterized, highly dynamic cinematography and unconventional visual grammar are not fully captured.
Future research may address these limitations by integrating text-to-3D generation for dynamic asset synthesis, enabling end-to-end motion generation models conditioned on narrative intent, and exploring fully generative camera control learned from large-scale cinematic data. Extending the framework beyond previsualization toward post-production rendering, automated editing, and multi-shot temporal optimization would further broaden its applicability.

\section{Application}

\begin{figure}[h]
    \centering
    \includegraphics[width=\linewidth]{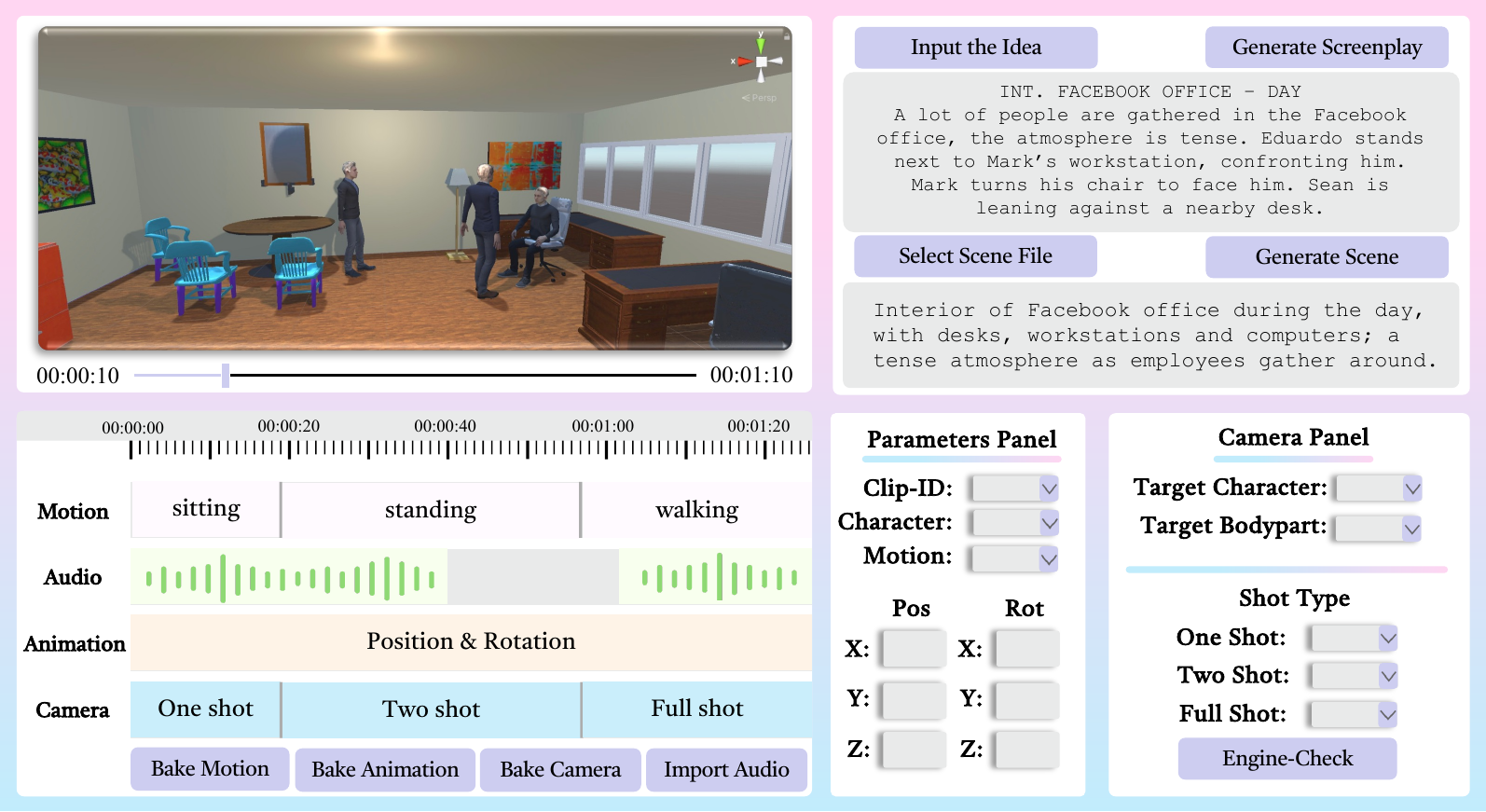}
    \caption{\textbf{Unity-Based Interface.}  
    Our system provides synchronized timeline tracks for characters and cameras, enabling real-time inspection, editing, and visualization across all stages.}
    \label{fig:unity_ui}
\end{figure}

We integrate \textbf{Mind-of-Director} into Unity as an interactive previz toolbox that supports human-in-the-loop authoring and real-time control as shown in Fig.~\ref{fig:unity_ui}. 
A key component is the Unity Timeline, where each character has synchronized tracks for motion, dialogue, and facial animation, while a dedicated camera track manages movement and framing.
Compared with traditional workflows, our framework shortens the previz process from several days to under 30 minutes, reducing human and software costs by over 90\%. It enables filmmakers to iterate rapidly, explore creative variations, and prototype ideas efficiently within an interactive environment.

% \vspace{-4pt}
% \begin{table}[h]
% \centering
% \footnotesize
% \caption{A brief comparison between \textbf{Traditional Previz} pipeline and \textbf{Mind-of-Director} framework on efficiency and cost for a 3-minute film previz sequence.}
% \vspace{-5pt}
% \begin{tabularx}{\linewidth}{
% >{\centering\arraybackslash}p{0.23\linewidth}
% >{\centering\arraybackslash}m{0.33\linewidth}
% >{\centering\arraybackslash}m{0.31\linewidth}
% }
% \toprule
% \textbf{Aspect} & \textbf{Traditional Previz} & \textbf{Mind-of-Director} \\
% \midrule
% \textbf{Human} & 6--10 specialists & 1 creator \\
% \textbf{Software} & Maya, Unreal, Houdini & Unity, integrated tools \\
% \textbf{Software Cost} & $\sim$\$5{,}000--\$10{,}000 & $\sim$\$50--\$100 \\
% \textbf{Time Cost} & 3--5 days & $\sim$25 min \\
% \textbf{Token Usage} & -- & 400K in / 90K out \\
% \bottomrule
% \end{tabularx}
% \label{tab:efficiency_comparison}
% \vspace{-8pt}
% \end{table}

\section{Conclusion}
We introduce Mind-of-Director, a multi-modal, multi-agent framework for automated film previsualization. By structuring collaborative reasoning across script development, scene design, character behaviour, and camera planning, the system demonstrates how distributed agent interaction can improve coherence and constraint satisfaction in complex creative pipelines.
Extensive quantitative and human evaluations validate the benefits of visual grounding, iterative collaboration, and engine-integrated reasoning. Beyond previsualization, this work provides a structured paradigm for organizing generative models into coordinated creative systems, offering a foundation for future research in AI-assisted cinematic production.

%% file: supp.tex
\begin{strip}
\centering
{\LARGE \textbf{Supplementary Material}}
\vspace{0.5em}
\end{strip}

\section*{Overview}
\label{sec:overview}
In this supplementary material, we present:
\begin{itemize}[topsep=0pt, itemsep=0em, parsep=0pt, partopsep=0pt, left=0em]
    \item Annotations of PrevizPro dataset (Sec.~\ref{sec:supp_dataset}).
    \item Algorithm for agent collaboration (Sec.~\ref{sec:supp_algorithm}).
    \item Detailed pipeline for Virtual Scene Design (Sec.~\ref{sec:supp_scene}).
    \item Examples for camera template library (Sec.~\ref{sec:supp_camera}).
    \item Human evaluation example (Sec.~\ref{sec:human_evaluation}).
    \item Example prompts for the pipeline (Sec.~\ref{sec:supp_prompts}).
    \item More qualitative results of our framework (Sec.~\ref{sec:supp_results}).
\end{itemize}

\section{PrevizPro Dataset Annotation}
\label{sec:supp_dataset}
{
\frenchspacing
The \textbf{PrevizPro} dataset provides detailed human annotations for evaluating motion and camera performance within our framework. All annotations are performed after our system predicts character states and positions, ensuring that expert decisions are grounded in spatial context. After dividing an act into dialogue-based clips, film experts annotate:
\begin{itemize}[topsep=0pt, itemsep=0em, left=0em]
    \item \textbf{Character Motions}:  
    For each clip, experts assign a motion from the motion library to every character based on the state and position, accompanied by a brief reason.
    \item \textbf{Camera Templates}:  
    For each clip, experts select the most appropriate camera template from the template library, also accompanied by a brief reason.
\end{itemize}
}

\vspace{3pt}
% ---------- Annotation: Character Motions ----------
\begin{table}[h]
\centering
\small
\captionof{table}{An annotation example for character motions.}
\label{tab:clip_actions}
\vspace{-5pt}
\begin{tabularx}{\linewidth}{
  >{\centering\arraybackslash}p{0.20\linewidth}
  >{\centering\arraybackslash}p{0.17\linewidth}
  >{\centering\arraybackslash}X
}
\toprule
\textbf{Character} & \textbf{State} & \textbf{Action} \\
\midrule
Clara & standing & ID 20: \textit{Thoughtful Head Shake} \\
Elias & sitting  & ID 29: \textit{Sitting Idle} \\
Maya  & standing & ID 17: \textit{Talking} \\
\midrule
\multicolumn{3}{@{}p{\linewidth}}{\textit{Reasons:} Clara reacts with restrained discomfort; Elias remains withdrawn; Maya speaks with subtle tension.} \\
\bottomrule
\end{tabularx}
\vspace{-8pt}
\end{table}

% ---------- Annotation: Camera Template ----------

\begin{table}[h]
\centering
\small
\captionof{table}{An annotation example for camera template.}
\label{tab:clip_shots}
\vspace{-5pt}
\begin{tabularx}{\linewidth}{
>{\centering\arraybackslash}p{0.23\linewidth}
>{\centering\arraybackslash}p{0.33\linewidth}
>{\centering\arraybackslash}X
}
\toprule
\textbf{Type} & \textbf{Specs} & \textbf{Subjects} \\
\midrule
two\_static &
\begin{tabular}{@{}l@{}}
Relation: distant \\
Framing: OTS\_pair \\
Shot size: MS \\
Angle: Eye
\end{tabular}
& Elias, Maya \\
\midrule
\multicolumn{3}{@{}p{\linewidth}}{\textit{Reason:} Maintains spatial logic (Elias at S1, Maya at T4), captures face-to-face interaction, and emphasizes the attention shift.}\\
\bottomrule
\end{tabularx}
\vspace{-8pt}
\end{table}

{
\frenchspacing
We provide an annotation example for motion in Table~\ref{tab:clip_actions} and an example for camera in Table~\ref{tab:clip_shots}. 
For camera evaluation in the main paper, we consider a prediction correct if the \textit{camera type} matches the expert annotation.
}
\newpage

\section{Agent Collaboration Algorithms}
\label{sec:supp_algorithm}

In the main paper, we describe how performing regions are refined through multi-agent collaboration among the \textit{Director}, \textit{Cinematographer}, \textit{Scene Designer}, and \textit{Actors}.  
We extend the \textit{Discuss–Revise–Judge} mechanism to a multi-role setting: a first director \(D_1\) proposes an initial blocking plan; the cinematographer \(P\) evaluates it from a shot-feasibility perspective; the scene designer \(SD\) evaluates spatial layout and collision risks; and each actor \(A_k\) evaluates their own performance in terms of character intent. All feedback is aggregated and revised by \(D_1\), and then a second director \(D_2\) for judgement as described in Algorithm~\ref{alg:multiagent_djr}.

\vspace{-4pt}
\begin{algorithm}[h]
\footnotesize
\caption{Discuss-Revise-Judge for Blocking}
\label{alg:multiagent_djr}
\begin{algorithmic}[1]
\State \textbf{Notation:}
\Statex \(\;C_{ij}\): the \(j\)-th clip in act \(A_i\)
\Statex \(\;B_{ij}^{(t)}\): blocking plan for \(C_{ij}\) at iteration \(t\)
\Statex \(\;D_1\): Director 1 (proposer) \quad \(D_2\): Director 2 (judge)
\Statex \(\;P\): Cinematographer \quad \(SD\): Scene Designer
\Statex \(A_k\): Actor for character \(k\)
\Statex \(F^{\text{cine}}_{ij}, F^{\text{scene}}_{ij}, F^{\text{actor}}_{ij,k}\): feedback from each role
\Statex \(\Phi(\cdot, \cdot)\): revision operator combining a plan and feedback
\vspace{4pt}

\For{each clip \(C_{ij}\)}
    \State \(D_1:\) initialize \(B_{ij}^{(0)}\)
    \For{\(t = 0\) to max\_rounds}
        \State \Comment{Parallel discussion and feedback}
        \State \(B_{ij}^{(t)} \xrightarrow{P} F^{\text{cine}}_{ij}\) \hfill \(\triangleright\) shot and visibility feedback
        \State \(B_{ij}^{(t)} \xrightarrow{SD} F^{\text{scene}}_{ij}\) \hfill \(\triangleright\) layout and collision feedback
        \For{each actor \(A_k\)}
            \State \(B_{ij}^{(t)} \xrightarrow{A_k} F^{\text{actor}}_{ij,k}\) \hfill \(\triangleright\) character-level performance feedback
        \EndFor
        \State Aggregate feedback: \(F_{ij} = \{F^{\text{cine}}_{ij}, F^{\text{scene}}_{ij}, F^{\text{actor}}_{ij,k}\}_k\)
        \State \(B_{ij}^{(t+1)} = \Phi(B_{ij}^{(t)}, F_{ij})\) \hfill \(\triangleright\) \(D_1\) revises blocking/behaviour
        \State \(B_{ij}^{(t+1)} \xrightarrow{D_2} \text{approve?}\)
        \If{approved}
            \State \(B_{ij}^{*} \leftarrow B_{ij}^{(t+1)}\)
            \State \textbf{break}
        \EndIf
    \EndFor
\EndFor
\State \textbf{Output:} refined behaviour plans \(\{B_{ij}^{*}\}_{i,j}\) for all clips
\end{algorithmic}
\end{algorithm}

\section{Virtual Scene Design Pipeline}
\label{sec:supp_scene}

In the main paper, Sec.~3.2 describes the \textit{Virtual Scene Design} module at a high level. 
Here we provide the full pipeline in Algorithm~\ref{alg:scene_pipeline}, which follows four stages:
(1) guidance image generation,
(2) object retrieval,
(3) base environment construction,
(4) ornament enhancement.
\newpage

\begin{algorithm*}[t]
\caption{Virtual Scene Design Pipeline}
\label{alg:scene_pipeline}
\begin{algorithmic}[1]
\State \textbf{Input:} scene description \(SD_i\) for act \(A_i\)
\State \textbf{Output:} 3D environment \(E_i\)

\vspace{6pt}
\State \textbf{Stage 1: Guidance Image Generation}
\State \(G_i \leftarrow \texttt{GenGuidanceImage}(SD_i)\) \Comment generate a 2D guidance image encoding composition and object layout

\vspace{4pt}
\State \textbf{Stage 2: Object Retrieval}
\State \(\texttt{detect\_img} \leftarrow \texttt{Florence2}(G_i)\) \Comment object detection on guidance image
\State \(\texttt{filtered\_detect} \leftarrow \texttt{Qwen3\_Filter}(\texttt{detect\_img})\) \Comment remove invalid / redundant boxes
\State \(\texttt{segments} \leftarrow \texttt{SAM}(\texttt{filtered\_detect})\) \Comment segment objects
\State \(\texttt{inpainted\_objs} \leftarrow \texttt{QwenImageEdit}(\texttt{segments})\) \Comment complete occluded or fragmented objects
\State \(\mathcal{A} \leftarrow \emptyset\) \Comment retrieved asset set
\For{each object crop \(o \in \texttt{inpainted\_objs}\)}
    \State \(\texttt{desc}_o \leftarrow \texttt{Qwen3\_Describe}(o)\) \Comment detailed textual description
    \State \(s_{\text{img}} \leftarrow \texttt{CLIP}(o, \texttt{desc}_o)\) \Comment image--text similarity
    \State \(s_{\text{text}} \leftarrow \texttt{SBERT}(\texttt{desc}_o, \text{asset\_names})\) \Comment text--text similarity to Objaverse names
    \State \(a^* \leftarrow \arg\max_{a \in \text{Objaverse}} (s_{\text{img}} + s_{\text{text}})\)
    \State \(\mathcal{A} \leftarrow \mathcal{A} \cup \{a^*\}\)
\EndFor

\vspace{6pt}
\State \textbf{Stage 3: Base Environment Construction}
\State \(\texttt{scene\_graph} \leftarrow \texttt{BuildSceneGraph}(\texttt{filtered\_detect}, \mathcal{A}, SD_i)\)
\State \(\mathcal{A}_{\text{anchor}}, \mathcal{A}_{\text{non}}\leftarrow\texttt{SplitAnchors}(\texttt{scene\_graph})\)
\State \(\texttt{grid\_floor} \leftarrow \texttt{InitGrid}(M, N)\) \Comment ground-level occupancy grid
\State \(\mathcal{G}_{\text{top}} \leftarrow \emptyset\) \Comment per-anchor top-surface grids

\For{each anchor \(a \in \mathcal{A}_{\text{anchor}}\)}
    \State \(\texttt{pose}_a \leftarrow \texttt{PlaceAnchor}(a, \texttt{scene\_graph}, \texttt{grid\_floor})\)
    \State \(\texttt{UpdateGrid}(\texttt{grid\_floor}, a, \texttt{pose}_a)\)
    \State \(\texttt{grid\_top}(a) \leftarrow \texttt{InitTopGrid}(a)\)
    \State \(\mathcal{G}_{\text{top}} \leftarrow \mathcal{G}_{\text{top}} \cup \{\texttt{grid\_top}(a)\}\)
\EndFor

\For{each non-anchor \(b \in \mathcal{A}_{\text{non}}\)}
    \State \(\texttt{target\_grid} \leftarrow \texttt{SelectGrid}(\texttt{grid\_floor}, \mathcal{G}_{\text{top}}, \texttt{scene\_graph}, b)\)
    \State \(\texttt{pose}_b \leftarrow \texttt{PlaceNonAnchor}(b, \texttt{scene\_graph}, \texttt{target\_grid})\)
    \State \(\texttt{UpdateGrid}(\texttt{target\_grid}, b, \texttt{pose}_b)\)
\EndFor

\vspace{4pt}
\State \textbf{Stage 4: Ornament Enhancement}
\State \(\texttt{empty\_cells} \leftarrow \texttt{GetEmptyCells}(\texttt{grid\_floor}, \mathcal{G}_{\text{top}})\)
\For{each cell \(c \in \texttt{empty\_cells}\)}
    \State \(\texttt{orn\_type} \leftarrow \texttt{LLM\_SuggestOrnament}(SD_i, c)\) \Comment e.g., plant, lamp, book
    \If{\texttt{CheckPlacementRules}(\texttt{orn\_type}, c, \texttt{scene\_graph})}
        \State \(\texttt{PlaceOrnament}(\texttt{orn\_type}, c)\)
        \State \(\texttt{UpdateGridAtCell}(c)\)
    \EndIf
\EndFor

\vspace{6pt}
\State Assemble all placed assets and ornaments into a 3D scene
\State \(\;E_i \leftarrow \texttt{BuildEnvironment}(\mathcal{A}, \texttt{scene\_graph}, \texttt{grid\_floor}, \mathcal{G}_{\text{top}})\)
\State \textbf{Return:} \(E_i\) as the final 3D virtual environment
\end{algorithmic}
\end{algorithm*}

\section{Camera Template Library}
\label{sec:supp_camera}

In the main paper (Sec.~3.4), we introduce a parameterized camera template library.  
Below we provide two representative templates in Examples~\ref{example:twostatic}--\ref{example:singledynamic}.

\begin{tcolorbox}[title=\textbf{Example 1: TWOSTATIC},
  coltitle=black, colback=white, colframe=black!10, arc=2mm, 
  boxrule=0pt, breakable, left=1mm, right=1mm]
\scriptsize
\refstepcounter{figure}\label{example:twostatic}
\begin{verbatim}
def TWOSTATIC(A, B, relation, framing, size, angle):
    """
    Intent: Present relationship balance or tension.
    Desc.: Two subjects in a stable, static layout.
    Use When: Dialogue, argument, reconciliation.

    Output:
    {
      "type": "two_static",
      "subjects": [A, B],
      "relation": relation,
      "framing": framing,
      "shot_size": size,
      "angle": angle,
      "rationale": "xxx"
    }

    RELATIONAL EXTENSIONS:
     relation: {"equal","dominant_A",
                "dominant_B", "intimate",...}
     framing: {"two_shot","OTS_pair",
               "profile_duet","split_depth",...}
     focus_mode: {"both","focus_A",
                  "focus_B","rack_between",...}
    """
\end{verbatim}
\end{tcolorbox}

\vspace{-15pt}
\begin{tcolorbox}[title=\textbf{Example 2: SINGLEDYNAMIC\_PEDESTAL},
  coltitle=black, colback=white, colframe=black!10, arc=2mm, 
  boxrule=0pt, breakable, left=1mm, right=1mm]
\scriptsize
\refstepcounter{figure}\label{example:singledynamic}
\begin{verbatim}
def SINGLEDYNAMIC_PEDESTAL(subject, 
                           start_elev, 
                           end_elev, ease):
    """
    Intent: Modify vertical dominance or presence.
    Description: Pedestal move around a single subject.
    Use When: Emotional change, stand/sit transitions.

    Output:
    {
      "type": "pedestal",
      "subject": subject,
      "start_elev": start_elev,
      "end_elev": end_elev,
      "ease": ease,
      "rationale": "xxx"
    }

    VALID PARAMETERS:
      start_elev, end_elev: {"low","eye","high","top"}
      ease: {"linear","ease_in", 
             "ease_out","ease_in_out"}
    """
\end{verbatim}
\end{tcolorbox}

\vspace{-8pt}
\section{Human Evaluation Examples}
\label{sec:human_evaluation}
\begin{figure*}[h]
    \centering
    \includegraphics[width=\linewidth]{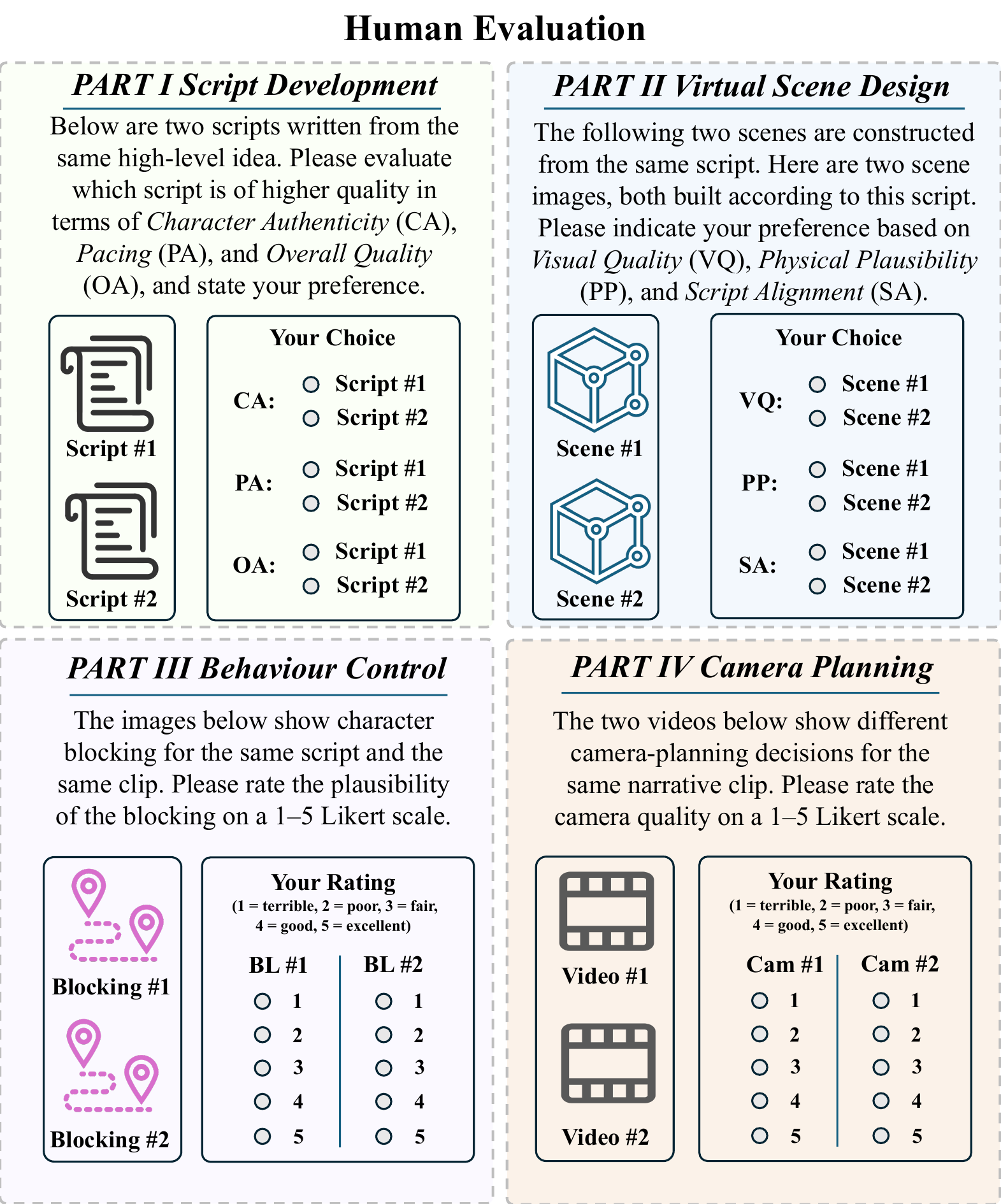}
    \caption{
    \textbf{Overview of Human Evaluation.}~
    The questionnaire consists of four parts:  
    \textbf{Part I} (Script Development) and \textbf{Part II} (Virtual Scene Design) are jointly rated by both experts and general users,  
    while \textbf{Part III} (Behaviour Control) and \textbf{Part IV} (Camera Planning) require domain expertise and are therefore evaluated exclusively by film experts.
    }
    \label{fig:human_evaluation}
\end{figure*}

Our human evaluation covers four components: \textit{script quality}, \textit{scene quality}, \textit{character blocking}, and \textit{camera planning}.  
To illustrate how expert and user feedback is collected, we provide a complete evaluation example for one screenplay.  
As shown in Figure~\ref{fig:human_evaluation}, the example includes detailed ratings and comments from both film-school experts and general users, demonstrating how human judgment complements our quantitative metrics across all modules.

\section{Prompt Design Examples}
\label{sec:supp_prompts}

To illustrate how multi-agent collaboration is instantiated in our framework, we provide a set of representative prompts for key stages. 
These examples highlight role-specific reasoning and structured interaction patterns, rather than exhaustively listing all prompts used in the system. 
Additional prompts are available in the project page.

\subsection{Script Development}

We present representative prompts for scene-level planning and multi-agent refinement.

\begin{tcolorbox}[title=\textbf{Scene Planning Prompt}, coltitle=black, colback=gray!10, colframe=black!10, arc=2mm, boxrule=0pt]
\footnotesize
\textbf{Task:} Given a film topic and character set, decompose the narrative into structured scenes.

\textbf{Key Requirements:}
\begin{itemize}[leftmargin=1em]
\item Define sub-topics for each scene
\item Select participating characters
\item Provide location, story plot, and dialogue goal
\end{itemize}

\textbf{Output:} Structured JSON describing scene decomposition.
\end{tcolorbox}

\vspace{-6pt}

\begin{tcolorbox}[title=\textbf{Director Revision Prompt}, coltitle=black, colback=gray!10, colframe=black!10, arc=2mm, boxrule=0pt]
\footnotesize
\textbf{Task:} Revise screenplay based on multi-agent feedback.

\textbf{Key Requirements:}
\begin{itemize}[leftmargin=1em]
\item Incorporate actor and director feedback
\item Maintain narrative consistency
\item Improve dramatic pacing and coherence
\end{itemize}

\textbf{Output:} Updated screenplay in structured JSON format.
\end{tcolorbox}

\subsection{Virtual Scene Design}

We include prompts for scene graph construction and spatial placement, which are central to enforcing structural consistency in generated environments.

\begin{tcolorbox}[title=\textbf{Scene Graph Construction}, coltitle=black, colback=gray!10, colframe=black!10, arc=2mm, boxrule=0pt]
\footnotesize
\textbf{Task:} Construct a structured scene graph from a detection image.

\textbf{Key Requirements:}
\begin{itemize}[leftmargin=1em]
\item Identify anchor and non-anchor objects
\item Infer spatial relations between objects
\item Assign placement rules for non-anchor objects
\end{itemize}

\textbf{Output:} JSON scene graph including objects and relations.
\end{tcolorbox}

\vspace{-6pt}

\begin{tcolorbox}[title=\textbf{Object Placement}, coltitle=black, colback=gray!10, colframe=black!10, arc=2mm, boxrule=0pt]
\footnotesize
\textbf{Task:} Place objects in a 3D environment following scene graph constraints.

\textbf{Key Requirements:}
\begin{itemize}[leftmargin=1em]
\item Place anchor objects first without collision
\item Place non-anchor objects using relational rules
\item Maintain spatial plausibility and layout consistency
\end{itemize}

\textbf{Output:} Structured JSON specifying object positions and orientations.
\end{tcolorbox}

\subsection{Character Behaviour Control}

We present prompts for initial blocking and director-level revision to illustrate the Discuss–Revise–Judge interaction.

\begin{tcolorbox}[title=\textbf{Initial Blocking Plan}, coltitle=black, colback=gray!10, colframe=black!10, arc=2mm, boxrule=0pt]
\footnotesize
\textbf{Task:} Generate an initial blocking plan for all characters.

\textbf{Key Requirements:}
\begin{itemize}[leftmargin=1em]
\item Assign positions and states for each character per clip
\item Maintain temporal continuity across clips
\item Avoid unnecessary movement
\end{itemize}

\textbf{Output:} Structured JSON describing character trajectories and states.
\end{tcolorbox}

\vspace{-6pt}

\begin{tcolorbox}[title=\textbf{Blocking Revision (Director)}, coltitle=black, colback=gray!10, colframe=black!10, arc=2mm, boxrule=0pt]
\footnotesize
\textbf{Task:} Refine blocking plan using multi-agent feedback.

\textbf{Key Requirements:}
\begin{itemize}[leftmargin=1em]
\item Resolve conflicts across cinematography, scene design, and acting
\item Preserve dramatic intent
\item Ensure spatial validity and continuity
\end{itemize}

\textbf{Output:} Updated blocking plan in structured JSON.
\end{tcolorbox}

\subsection{Camera Planning}

We provide the final director synthesis prompt, which integrates multiple proposals into a coherent camera plan.

\begin{tcolorbox}[title=\textbf{Camera Plan Synthesis}, coltitle=black, colback=gray!10, colframe=black!10, arc=2mm, boxrule=0pt]
\footnotesize
\textbf{Task:} Generate the final camera plan by synthesizing multiple cinematographer proposals.

\textbf{Key Requirements:}
\begin{itemize}[leftmargin=1em]
\item Ensure physical validity and consistency with character motion
\item Maintain cinematic clarity and composition
\item Apply minimal necessary modifications
\end{itemize}

\textbf{Output:} Final camera plan in structured JSON format.
\end{tcolorbox}

\section{Additional Qualitative Results}
\label{sec:supp_results}

\begin{figure*}[h]
    \centering
    \includegraphics[width=\linewidth]{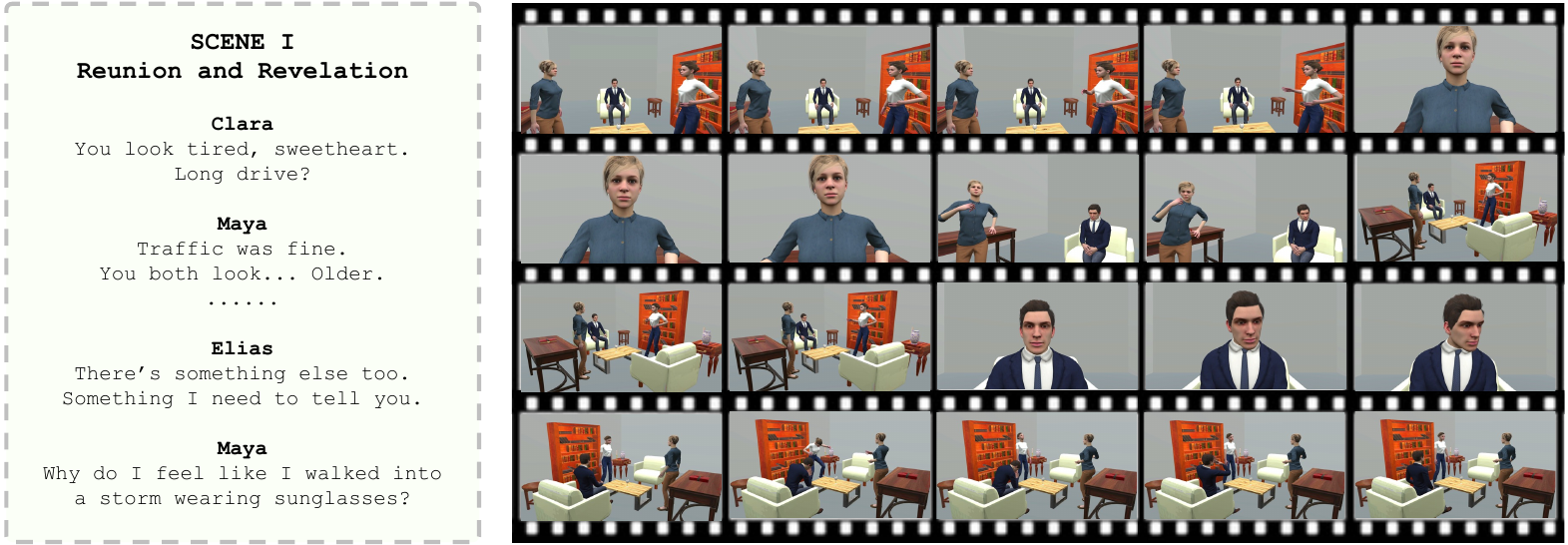}
    \caption{
        The previs sequences of the original screenplay \textit{The Last Supper}, \textit{Scene I}. 
    }
    \label{fig:s1}
\end{figure*}

\begin{figure*}[h]
    \centering
    \includegraphics[width=\linewidth]{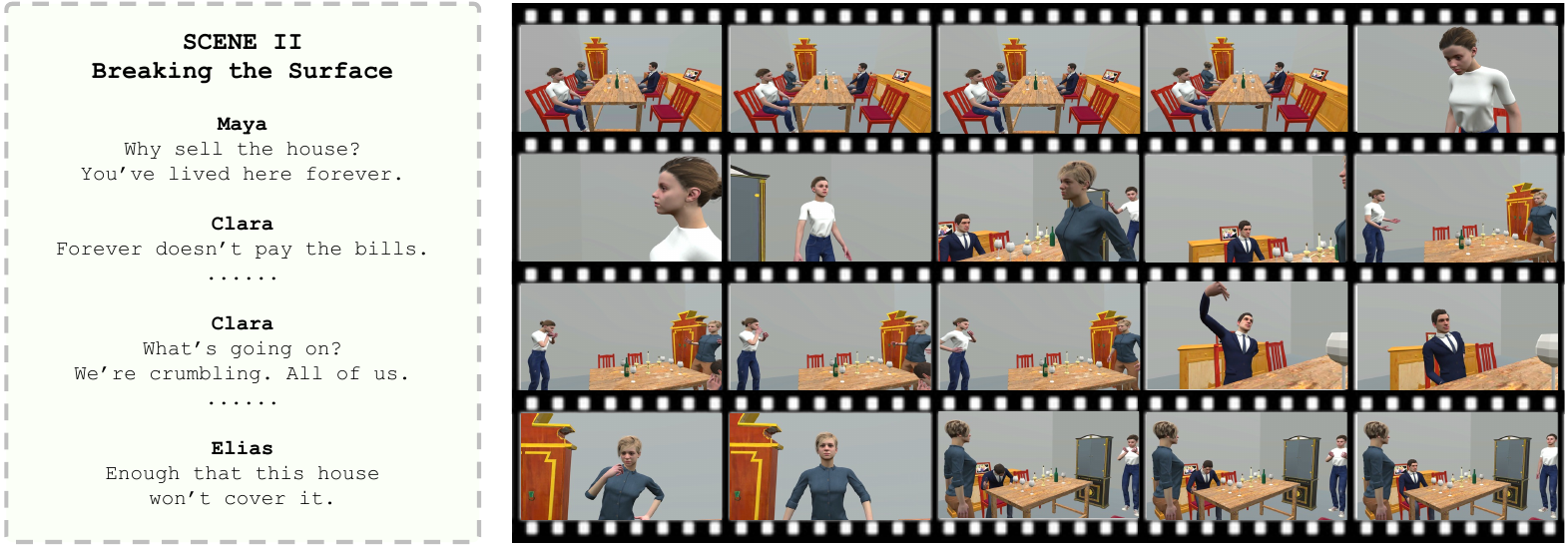}
    \caption{
        The previs sequences of the original screenplay \textit{The Last Supper}, \textit{Scene II}. 
    }
    \label{fig:s2}
\end{figure*}

\begin{figure*}[h]
    \centering
    \includegraphics[width=\linewidth]{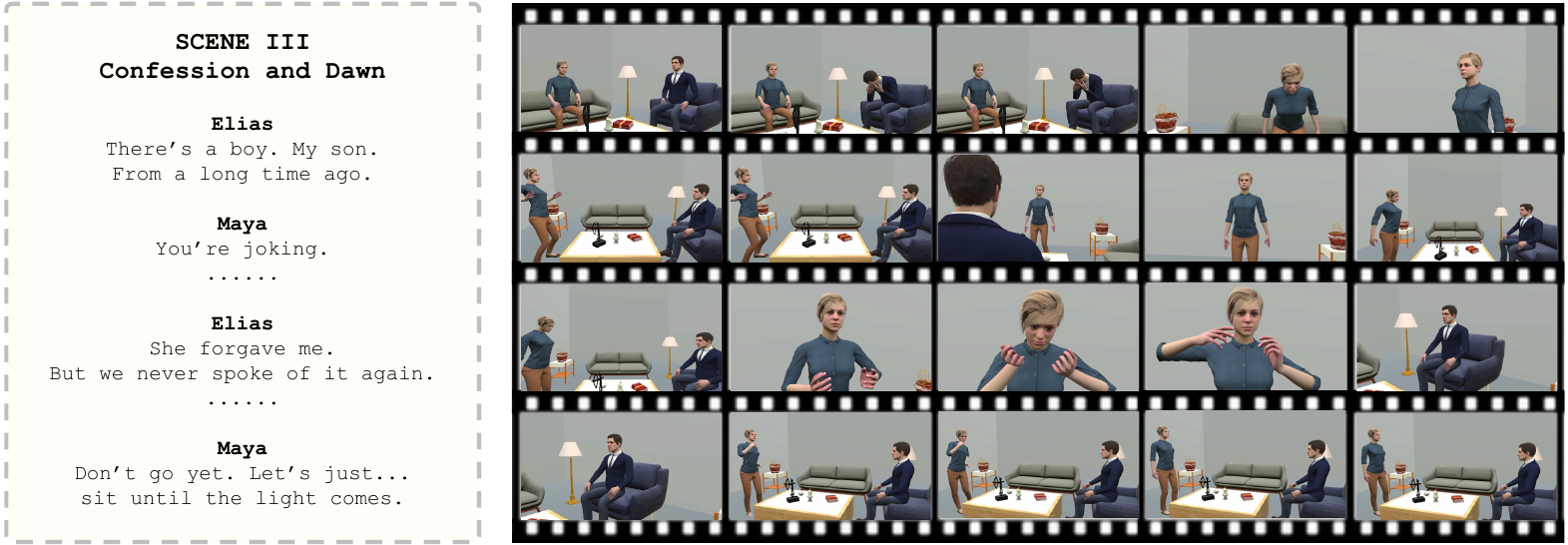}
    \caption{
        The previs sequences of the original screenplay \textit{The Last Supper}, \textit{Scene III}. 
    }
    \label{fig:s3}
\end{figure*}

\begin{figure*}[h]
    \centering
    \includegraphics[width=\linewidth]{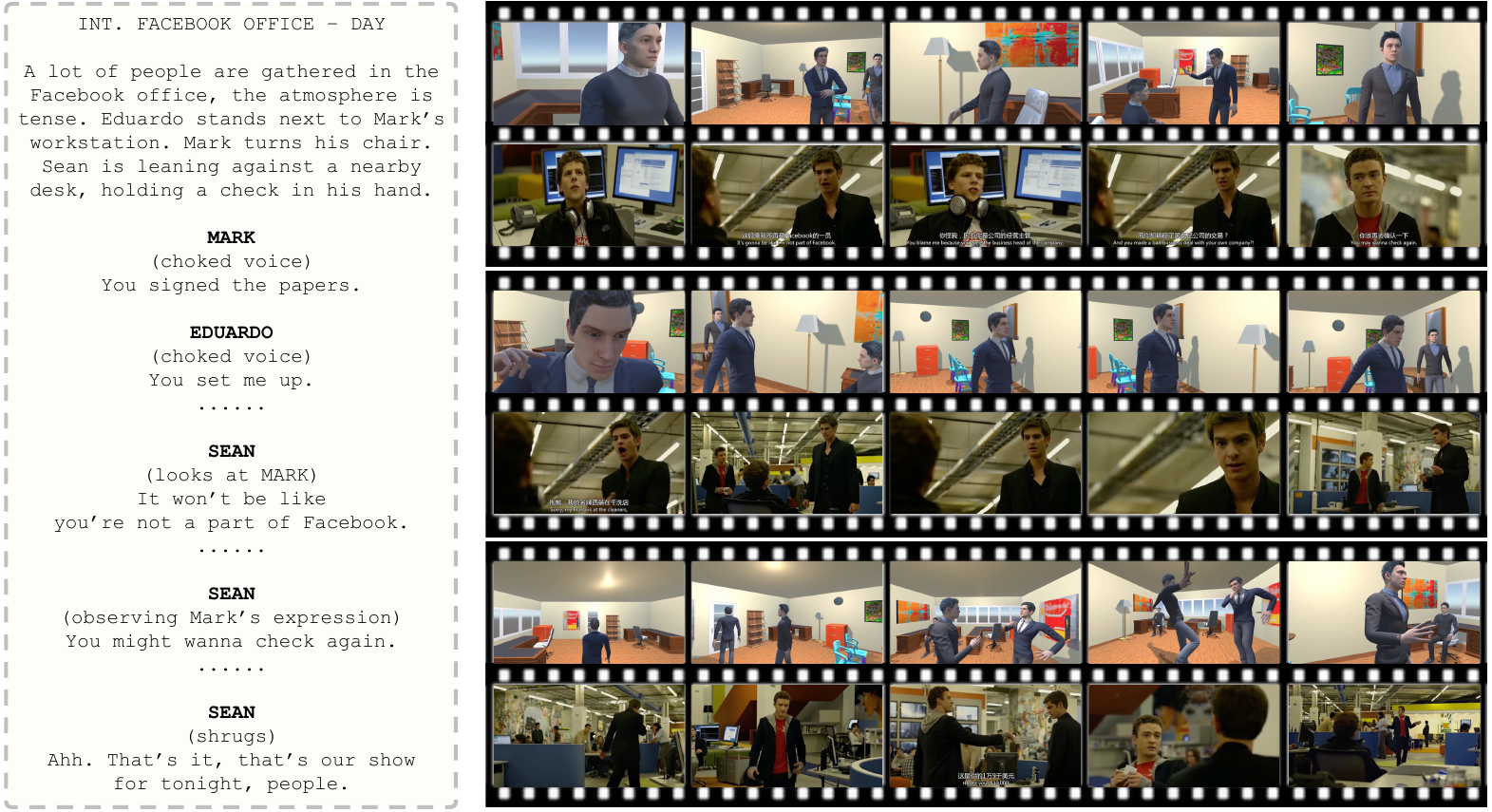}
    \caption{
        The previs sequences of the movie \textit{The Social Network} directed by \textit{David Fincher} in $2010$. 
    }
    \label{fig:social_network}
\end{figure*}

\begin{figure*}[h]
    \centering
    \includegraphics[width=\linewidth]{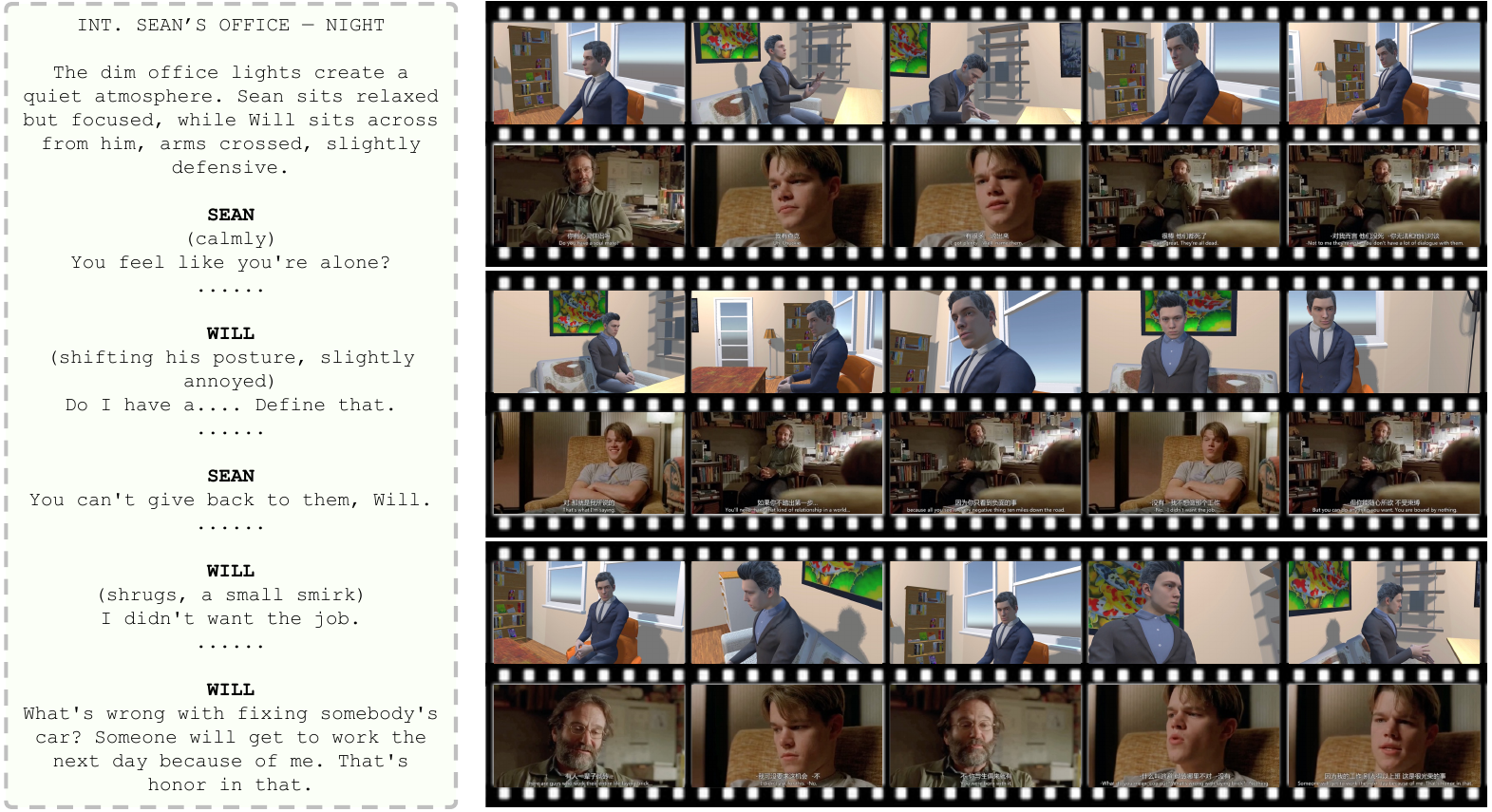}
    \caption{
        The previs sequences of the movie \textit{Good Will Hunting} directed by \textit{Gus Van Sant} in $1997$.
    }
    \label{fig:good_will_hunting}
\end{figure*}

In Figures~\ref{fig:s1}--\ref{fig:s3}, we showcase a complete previz sequence produced directly from a high-level idea \textit{The Last Supper}, illustrating how our framework transforms an idea into final previz sequences.  
To further demonstrate practical applicability, we additionally include two examples derived from real movie scripts in Figures~\ref{fig:social_network}--\ref{fig:good_will_hunting}, \textit{The Social Network} directed by \textit{David Fincher} in 2010 and \textit{Good Will Hunting} directed by \textit{Gus Van Sant} in 1997.   
These results highlight the versatility of \textbf{Mind-of-Director} in handling both open-ended creative ideation and professionally written screenplays across diverse narrative contexts.

